\documentclass[final]{siamltex}
\usepackage{algorithm,algpseudocode}
\usepackage{latexsym, graphicx, epsfig, amsmath, amsfonts,amssymb,bm}
\usepackage{caption, subcaption}
\usepackage{epstopdf}
\usepackage{booktabs}
\usepackage{array}
\usepackage{color}
\usepackage{url}
\usepackage{comment}
\usepackage{diagbox}

\usepackage{tikz} 
\usetikzlibrary{arrows.meta, positioning}
\usetikzlibrary{positioning}
\usepackage{subcaption}

\def\Dt{\Delta t}

\allowdisplaybreaks

\def\be{\begin{equation}}
\def\ee{\end{equation}}

\def\R{{\mathbb R}}

\def\H{\mathcal{H}}

\def\0{\mathbf{0}}

\title{Targeted Digital Twin via Flow Map Learning and Its Application to Fluid Dynamics}

\author{Qifan Chen\and Zhongshu Xu\and Jinjin Zhang\and Dongbin Xiu\thanks{Department of Mathematics,
		The Ohio State University, Columbus, OH 43210, USA. Emails:
		{\tt chen.11010, xu.4202, zhang.14647, xiu.16@osu.edu.} Funding: This 
		work was partially supported by AFOSR FA9550-24-1-0237.}
				}

\begin{document}
\maketitle
\begin{abstract}
We present a numerical framework for constructing a targeted digital twin (tDT) that directly models the dynamics of quantities of interest (QoIs) in a full digital twin (DT). The proposed approach employs memory-based flow map learning (FML) to develop a data-driven model of the QoIs using short bursts of trajectory data generated through repeated executions of the full DT. This renders the construction of the FML-based tDT an entirely offline computational process. During online simulation, the learned tDT can efficiently predict and analyze the long-term dynamics of the QoIs without requiring simulations of the full DT system, thereby achieving substantial computational savings.
After introducing the general numerical procedure, we demonstrate the construction and predictive capability of the tDT in a computational fluid dynamics (CFD) example: two-dimensional incompressible flow past a cylinder. The QoIs in this problem are the hydrodynamic forces exerted on the cylinder. The resulting tDTs are compact dynamical systems that evolve these forces without explicit knowledge of the underlying flow field. Numerical results show that the tDTs yield accurate long-term predictions of the forces while entirely bypassing full flow simulations.
\end{abstract}
\begin{keywords}
Targeted digital twin, flow map learning, data-driven modeling, computational fluid dynamics.
\end{keywords}

\section{Introduction} \label{sec:intro}

Digital Twin (DT) is a concept that has emerged in several industrial contexts over the past decades. Formally conceptualized by Michael Grieves and John Vickers in the 2000s, DT has since attracted significant attention from academia. Widely recognized as a paradigm shift in digital transformation, Digital Twins hold tremendous promise for revolutionizing industries and accelerating both decision-making and scientific discovery. A simple online search yields thousands of publications discussing various aspects of Digital Twins. We refer the readers to \cite{Grieves2023} and the references therein for an account of the history of Digital Twins, as well as \cite{NAE-DT2024, Crespi-2023,Iranshahi25} for comprehensive reviews of the current landscape.

A DT is a virtual representation of a physical system that undergoes continuous updates from real-time data, enabling simulation, analysis, control, and decision-making. A critical feature of DT is the bi-directional data exchange between the physical and digital twins. On the one hand, data from the physical twin are continuously incorporated into the digital twin, ensuring that the digital representation remains accurate with respect to the physical twin and its environment. On the other hand, the digital twin provides real-time control adjustments and decisions to the physical twin in response to environmental changes and operational objectives. In \cite{XiuTartakovsky_PDDP}, the former process was termed P2D (physical-to-digital), the latter D2P (digital-to-physical), and the overall bi-directional exchange was described as the PDP (physical-to-digital-to-physical) framework.

Although widely adopted, real-time implementation of this framework remains challenging due to its high computational cost. At the core of the DT framework lies a complete numerical solver for modeling the states of the physical twin. Also called the forward (problem) solver, this component provides accurate predictions and analyses of the physical system. However, when the physical twin is complex, which is almost always the case, the forward solver often requires extraordinarily sophisticated simulation software and incurs significant computational expense. The difficulty is compounded by the fact that many tasks in the P2D and D2P processes (e.g., optimization, control, parameter estimation, data assimilation, and uncertainty quantification) require repeated execution of the forward solver. This makes real-time DT implementation extremely challenging. Although recent works in the scientific computing community have sought to mitigate these challenges through various numerical strategies, cf. \cite{Farhat2022, Farhat2025,Wilcox_cmame24,Wilcox_jmd22,Wilcox_ncs21,Wilcox_ncs24,Rochinha21,Sharma22}, the fundamental difficulties remain.

\begin{figure}[htbp]
\begin{center}
\includegraphics[width=0.6\textwidth]{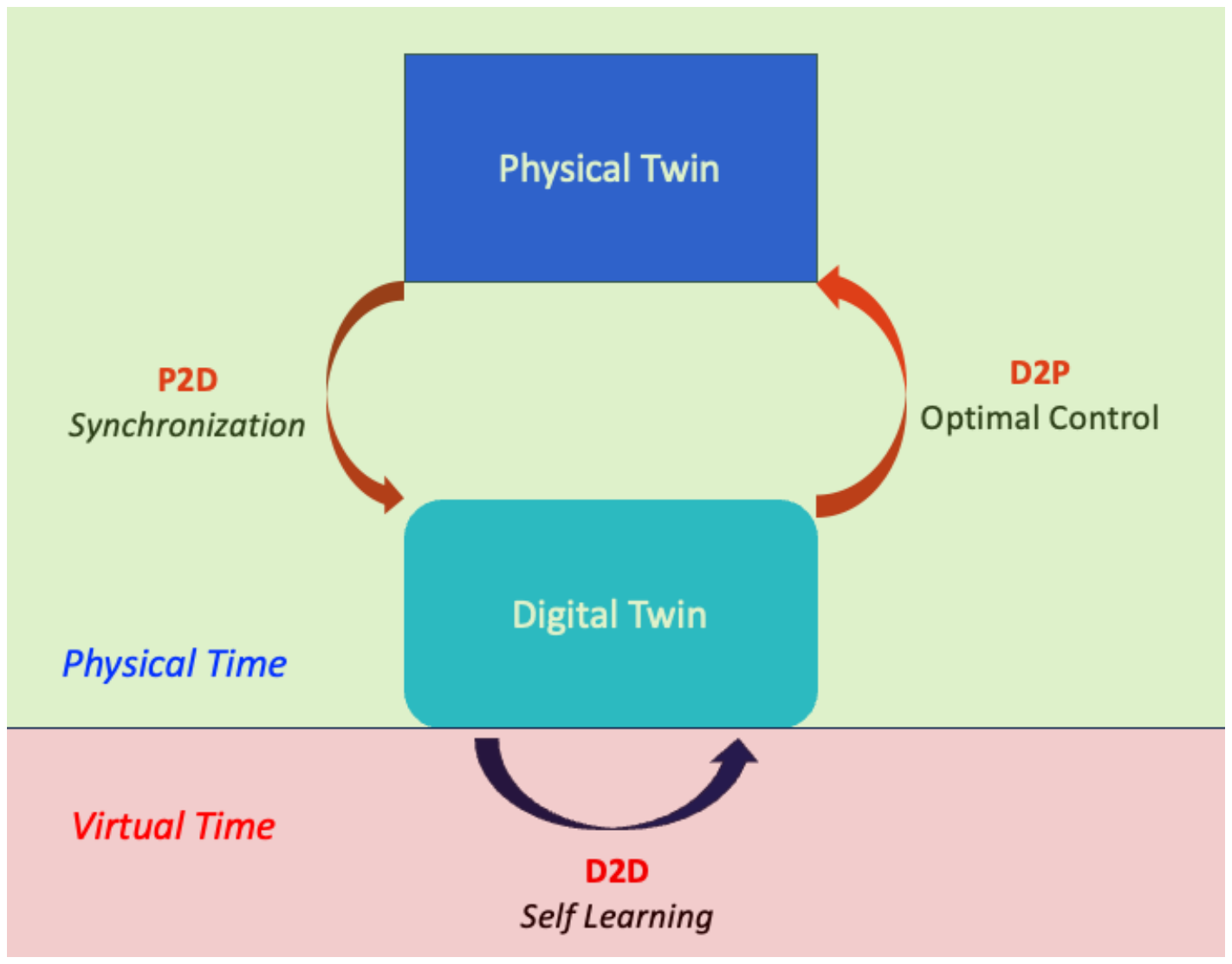}
\caption{PDDP (physical-to-digital-digital-to-physical) data flow for Digital Twin (\cite{XiuTartakovsky_PDDP}).}
\label{fig:PDDP}
\end{center}
\end{figure}

The purpose of this paper is to formalize the concept of the Targeted Digital Twin (tDT) and introduce a numerical strategy for its construction. The idea of tDT was first introduced in \cite{XiuTartakovsky_PDDP} within the PDDP (physical-to-digital-digital-to-physical) framework. Compared to the standard PDP bi-directional data flow framework, the PDDP framework introduces an additional D2D (digital-to-digital) loop (see Figure \ref{fig:PDDP}). The D2D process occurs entirely in virtual time and does not require real data or inputs from the physical twin. As a self-learning process, one of its main objectives is to construct a targeted digital twin for the full digital twin. A tDT models the dynamics of critical quantities of interest (QoIs) of the full digital twin. While the full DT often relies on a forward solver with an exceedingly large number of degrees of freedom, the key QoIs necessary for control and decision-making typically involve only a small number of quantities. For instance, decisions may depend on maximum stress, average temperature, or pressure forces at specific critical locations, rather than on the full state of all variables at every grid point. If a tDT, defined as a dynamical model for these QoIs, can be constructed without solving the full system, it would enable much faster model simulations for the QoIs and be suitable for real-time control and decision making of the DT based on the QoIs. This is the motivation for constructing tDTs within the PDDP framework (\cite{XiuTartakovsky_PDDP}).

QoIs are, however, defined as functions of the state variables, and their evaluation requires solving the complete system of the full digital twin. Mathematical models for QoIs that bypass the state variables do not exist. (In other words, if such a closed-form model for the QoIs were available, it would already be incorporated into the full DT.) Consequently, construction of a dynamical model for the QoIs must be data-driven, relying on training data generated from the full DT. In this paper, we demonstrate how this can be accomplished through the Flow Map Learning (FML) methodology. FML was first introduced in \cite{qin2018data} to model unknown autonomous dynamical systems, and later extended to parameterized systems \cite{QinChenJakemanXiu_IJUQ}, non-autonomous systems \cite{QinChenJakemanXiu_SISC}, PDEs \cite{chen2022deep,WuXiu_modalPDE}, and stochastic dynamical systems \cite{ChenXiu_JCP24, XuEtal_JMLMC24}. A notable extension addressed partially observed systems \cite{FuChangXiu_JMLMC20}, where memory-based FML was developed. This approach has proven accurate and effective for partially observed chaotic systems \cite{Churchill_pPDE23} and PDE systems \cite{ChurchillXiu_chaos22}. For a detailed review of FML, see \cite{ChurchillXiu_FML2023}.

In this paper, we present a computational procedure that adapts the memory-based FML approach for constructing a targeted DT from a full DT. We illustrate the method using a computational fluid dynamics (CFD) problem: two-dimensional incompressible flow past a cylinder, which is a well-studied benchmark. The forward solver for the full DT involves approximately $150{,}000$ degrees of freedom, which, while not exceedingly large, is nonetheless nontrivial. The QoIs are defined as the total hydrodynamic forces exerted on the cylinder, namely drag (the streamline component) and lift (the cross-flow component), yielding only two degrees of freedom. By conducting repeated simulations of the full DT under randomized conditions, we collect a sufficient number of short-burst training trajectories for the QoIs. A memory-based FML model is then constructed as the tDT for the forces. This tDT is an exceedingly compact model, with only two outputs: drag and lift. For prediction, a time series of cylinder forces is first acquired and used as the “initial condition” to synchronize the tDT with the full DT. Thereafter, the FML-based tDT can generate long-term predictions of the cylinder forces without any additional information from the full DT. That is, the tDT can analyze and predict the dynamical behavior of the flow-induced forces without explicit knowledge of the flow field. In a separate study, we consider the pressure distribution along the cylinder surface as the QoI and demonstrate that a tDT can also be constructed to accurately predict this distribution over long time horizons, again without requiring simulations of the full flow field.

This paper is organized as follows. Section \ref{sec:setup} provides
the formal definition of the targeted DT. Section \ref{sec:method}
details the computational procedure for constructing tDTs. Section
\ref{sec:examples} presents the CFD example and demonstrates the
effectiveness of the FML-based tDT approach. Conclusions are given in
Section \ref{sec:conclusions}.

\section{Targeted Digital Twin} \label{sec:setup}

In this section, we introduce the problem setup and the concept of
Targeted Digital Twin (tDT).

\subsection{Full Digital Twin}

Without being specific, we use the following general formula to denote the mathematical model of a complete full digital twin
\be \label{model}
u_t=\mathcal{F}(u; \alpha),
\ee
where $u$ represents the state variables, $\alpha$ the system parameters, and $\mathcal{F}$ the operator governing the evolution of the states. The system parameters $\alpha$ broadly include all the physical or hyperparameters in the model, as well as the geometry and boundary conditions.
Note that $\mathcal{F}$ is typically highly nonlinear and involves complex interactions among many (nonlinear) sub-systems of different physics and scales. We make a general assumption that the system \eqref{model} is well posed.

Since practically all mathematical models are solved numerically over discrete spatial and time domains, we shall adopt a discrete setting and consider, hereafter, without loss of much generality, time instances over a constant time step $\Delta t>0$,
$$
t_n = n\Delta t, \qquad n=0,1,\dots,
$$
and the full DT as a discrete numerical model
\be \label{DT}
U_{n+1} =F(U_n; p), \qquad U\in\R^{N_U}, \quad p\in\R^{N_p},
\ee
where $U_n$ represents the numerical discretization of the state variables $u$ over the spatial domain at time $t_n$, and $p$ represents all the parameters. These parameters $p$ include not only the system parameter $\alpha$ in the original mathematical model \eqref{model}, but also hyperparameters used in the construction of the discrete model \eqref{DT}. (Only variable parameters are considered here. Parameters with fixed values are treated as constants and not included in $p$.)
Typically, $U$ are the grid values of the state variables $u$
at a set of grid points when a collocation-type numerical method is
used, e.g., finite difference, or the expansion coefficients of $u$
over a set of basis functions when a projection-type numerical method
is used, e.g., the Galerkin method. To fully resolve the system \eqref{model},
the dimensionality of $U$, $N_U$, is usually very high. For example, it is not uncommon for a complex system solver to have $\sim 10^6$ mesh
points or more.

\subsection{Quantity-of-Interest}

For the model \eqref{model}, we consider its quantities-of-interest (QoIs)
\be \label{v}
v = \H(u;\beta),
\ee
where $\beta$ are the variable parameters associated with the definition of the QoI. These are the quantities that are of critical significance to the overall
performance or decision-making of the system
\eqref{model}. Typically, the dimension of QoI is low, often as low as
$O(1)$.

Correspondingly, for the discrete DT \eqref{DT}, we define the discrete approximation of the QoIs \eqref{v} as
\be \label{QoI}
V = H(U; q),
\ee
where $q$ represents all the variable (hyper) parameters used in the
construction of the operator $H$. 

We remark that a mathematical model of the dynamics of the QoIs \eqref{v} does not exist. To
evaluate the QoIs, one needs to first solve the system \eqref{model}, via
its discrete model \eqref{DT}, and then compute the QoIs through \eqref{QoI}. The dynamics of the QoIs thus follow
\be \label{V_n}
V_{n+1} = H(U_{n+1};q) = H\circ F(U_n; p, q).
\ee
Since the operators $H$ and $F$ in general do not commute, this does
not render a direct dynamical relationship between $V_{n+1}$ and $V_n$.

\subsection{Targeted Digital Twin}

The objective of this paper is to establish a numerical procedure for
constructing a dynamical model of the QoIs, $V$, that does not explicitly
depend on the full system state $U$. That is, we wish to create a
model in the form of
\be \label{tDT_idea}
V_{n+1} = G(V_n, \cdots),
\ee
where $G$ is to be determined and may involve other inputs. However,
the important feature is that $G$ does not rely on $U_n$. Hereafter we
shall call this model {\em targeted digital twin} (tDT) of the full DT \eqref{DT}.
More specifically, we employ the following loosely defined definition.
\begin{definition} \label{def:tDT}
A targeted digital twin is a numerical representation of the quantities of interest of a full digital twin, synchronized and dynamically updated with data from the
full digital twin.
\end{definition}

In other words, tDT is ``a digital twin of a digital twin".  The bi-directional data
exchange is exclusively between the tDT and the full DT and does not
involve the physical twin. This ensures that the construction,
learning, and updating of the tDT can be performed in virtual time by
executing the full DT model. This falls into the PDDP framework
proposed in \cite{XiuTartakovsky_PDDP}.
Note that this does not necessarily
exclude the use of real data from the physical twin, when such data
become available.

\subsection{Preliminary: Flow Map Learning}

The proposed numerical method for creating tDT has its foundation on a recently
developed methodology: flow map learning (FML). FML is a data driven
method for modeling unknown dynamics, first proposed in
\cite{qin2018data} for learning autonomous dynamical systems.
For an unknown autonomous system,
$x_t = f(x)$, $x\in\R^d$, where the governing equation $f:\R^d\to
\R^d$ is not known, FML seeks to learn its flow map operator
$\Phi:\R^d\to\R^d$ numerically, as opposed to learning $f$. The flow
map operator governs the evolution of the solutions, i.e., $x(t_n) =
\Phi_{t_n-t_s}(x(t_s))$. Therefore, when data of the solution
trajectories are available, we split the trajectory data into pairs of
data separated by one time step $\Delta t$. These pairs of data then
satisfy
$x(t+\Delta t) = \Phi_{\Delta t}(x(t))$, where $t$ is the ``starting time'' of each data pair.
In FML, we then seek a function $N:\R^d\to\R^d$ to minimize the following loss function
$$
\sum_{j=1}^{N_{data}} \left\| x^{(j)}(\Dt) - N(x^{(j)}(0))\right\|^2,
$$
where $N_{data}$ is the total number of such data pairs $\left(x^{(j)}(0), x^{(j)}(\Dt)\right)$, $j=1,\dots,N_{data}$. Note that the ``starting time'' of each pair is set to 0 as in autonomous systems, only the relative time matters.
Upon minimizing this loss function, we obtain $N\approx \Phi$ and a
predictive FML model
\be \label{fml0}
x_{n+1} = N(x_n),
\ee
which can be used to model the system dynamics when an initial
condition $x_0$ is specified.

A notable extension of FML is modeling an incomplete system (\cite{FuChangXiu_JMLMC20}).
Let $x = (z, w)$, where
$z\in\R^m$ are the observables and $w\in\R^{d-m}$ the missing variables. When trajectory data are available only
on $z$, which is a subset of the entire state variable $x$, the work
of \cite{FuChangXiu_JMLMC20}  established that modeling the dynamics of $z$ can be accomplished without any information of the missing variable $w$. Motivated by the celebrated Mori-Zwanzig formulation  (\cite{mori1965, zwanzig1973}),
\cite{FuChangXiu_JMLMC20} developed a discrete approximate Mori-Zwanzig formulation and established that a discrete dynamical model of $z$ exists in the following form,
\be \label{fml}
      z_{n+1} =
      N(z_n,z_{n-1}, \dots,z_{n-n_M}),  \qquad n\geq n_M,
 \ee
where $n_M\geq 0$ is the number of memory terms. In this case, $N:\R^{m\times (n_M+1)}\to\R^m$ is the flow map with memory $n_M$.
The special case of $n_M=0$ corresponds to the learning of an autonomous system when all the state variables $x$ are observed \eqref{fml0}.
The memory-based FML structure \eqref{fml} has been shown to be able to model a
wide class of partially observed systems. For a
review of FML modeling of unknown dynamical systems, see \cite{ChurchillXiu_FML2023}.


%

\section{Targeted Digital Twin Construction} \label{sec:method}

In this section, we first present the mathematical justification for
using the memory-based FML formulation \eqref{fml} to formulate tDT. We then
describe the general numerical procedure for constructing 
a tDT from a full DT.

\subsection{Mathematical Motivation}

To present the mathematical motivation for the proposed numerical
method, we return to the continuous setting of the problem. That is,
we consider the full mathematical model \eqref{model}, along with its
QoI \eqref{v}. 
By assuming $\H$ to be continuously differentiable with respect to $u$, we obtain
\be
v_t = \nabla_u \H \cdot u_t = \nabla_u\H \cdot \mathcal{F}(u; \beta).
\ee
Let $z=(u, v, \alpha, \beta)^T$ represent all the ``variables''. We formally obtain a
system of equations
\be
z_t =\left(
\begin{array}{c}
  u_t \\ v_t \\ \alpha_t \\ \beta_t
\end{array}
\right)
=\left(
  \begin{array}{c}
    \mathcal{F}(u; \alpha) \\
    \nabla_u\H\cdot\mathcal{F}(u; \beta) \\
    0 \\
    0
  \end{array}
  \right).
\ee
Note that we consider all the parameters $\alpha$ and $\beta$ as time independent. This is not a restricting condition. Time-dependent parameters are usually in the form of external signals, excitations, or controls. In this case, they are given inputs and do not need to be modeled. On the other hand, if a time-dependent parameter is not an external input but follows a certain intrinsic physical law, it should be included in the state variable $u$.

It is straightforward to see that the QoIs $v$ can now be considered a
subset of the variables in the full and larger system for $z$. The
Mori-Zwanzig motivated FML thus becomes applicable. When sufficient data of
$v$ are available, it is possible to learn a dynamical model for $v$
that does not depend explicitly on the knowledge $u$, $\alpha$, and $\beta$.

\subsection{Numerical Procedure for tDT} \label{sec:DT}

For numerical construction of tDT, we employ the discrete setting for the DT \eqref{DT} and the QoIs \eqref{QoI}. The dynamics of the QoI thus follow
\be \label{fullDT}
\left\{
\begin{aligned} 
  U_{n+1} &=F(U_n; p), \\
  V_{n+1} &= H(U_{n+1}; q).
\end{aligned}
\right.
\ee
The general procedure of constructing tDT consists of three main steps: (1)
repetitively execution of the system \eqref{fullDT} to generate the QoIs training
data; (2) learning and construction of the tDT model for $V$ in the form of \eqref{fml}; and (3)
synchronization of the tDT with the full DT to facilitate the
prediction of the QoIs $V$ without using the full DT.

For all the parameters $(p, q)$ involved in the full DT, we group them into two subsets, 
$
\{p,q\} =\{\gamma, \hat{\gamma}\},
$
where
\be \label{gamma}
\gamma = \{\textrm{explicit parameters}\}; \quad \hat{\gamma} = \{\textrm{hidden parameters}\}.
\ee
By ``explicit parameters" we refer to the parameters on which we have access to sufficient data and wish to study the QoIs' dependence. All other parameters are categorized as ``hidden parameters", for which we do not need any information or data. Since the hidden parameters do not appear anywhere during the study, we shall ignore them and freely call ``explicit parameters" simply ``parameters", unless confusion arises otherwise.
A tDT model for the QoIs takes the form
\be \label{tDT}
V_{n+1} = G(V_n, V_{n-1}, \dots, V_{n-n_M}; \gamma),
\ee
where $n_M\geq 1$ is the memory step. This follows the form of the memory-based FML model \eqref{fml} and involves $(n_M+2)$ consecutive entries of the QoIs $V$, 1 on the left-hand-side and $(n_M+1)$ on the right-hand-side.

\subsubsection{Full DT Model Simulation} \label{sec:average}

The first step of constructing tDT is to perform repeated full DT simulations to acquire training data for the tDT model \eqref{tDT}.

\begin{enumerate}
    \item{\emph{Set Up:}} Let $I_U$ be the domain-of-interest for the state variables,  and $I_P$ and $I_Q$ be the domain-of-interest for the parameters $p$ and $q$ in the full DT \eqref{fullDT}. Choose $N_{sim}\geq 1$ to be the number of full DT model simulations and 
    $N_{step}\geq (n_M+2)$ the total number of time steps of each simulation. 
    \item{\emph{Sampling:}} Conduct $N_{sim}$ samples of initial conditions and parameters in the corresponding domain-of-interest, i.e., sample 
    \be \label{rand}
    \left\{U_0^{(j)}\right\}_{j=1}^{N_{sim}}\subset I_U; \quad
    \left\{p^{(j)}\right\}_{j=1}^{N_{sim}}\subset I_P, \quad \left\{q^{(j)}\right\}_{j=1}^{N_{sim}}\subset I_Q.
    \ee
    In most cases, $I_U$, $I_P$, and $I_Q$ can be set as bounded domains. We advocate the use of random sampling with uniform distribution. However, other sampling strategies can be adopted based on the problem's needs. 
    \item{\emph{DT Simulations:}} Conduct $N_{sim}$ full DT simulations \eqref{fullDT}. Specifically, for each $j=1,\dots, N_{sim}$,
    \begin{itemize}
        \item Set the initial condition as $U_0^{(j)}$ and the parameters as $p^{(j)}$ and $q^{(j)}$;
        \item Simulate the full DT model \eqref{fullDT} for $N_{step}$ forward time steps;
        \item Evaluate the $(N_{step}+1)$ time history of the QoIs.
    \end{itemize}
    \item{\emph{Collect QoI Dataset:}} Discard all the state variables $U$ and the hidden parameters. Record only the time series data of the QoIs and their corresponding explicit parameters $\gamma$ to obtain the following QoI dataset
        \be \label{traj}
        \mathcal{Q} \triangleq \left\{\left. V_0^{(j)}, \dots, V_{N_{step}}^{(j)}; \gamma^{(j)}\right|j=1,\dots, N_{sim}\right\}.
        \ee
   
\end{enumerate}

The full DT model simulations are now completed. This is the most computationally expensive part of the process, where the full DT model is executed for a total number of $(N_{sim}\times N_{step})$ time steps. However, we remark that this cost is measured in terms of the number of time steps and not the number of full-scale complete model simulations. In standard scientific computing tasks, a full-scale complete model simulation is usually a long-term simulation lasting millions of time steps. Here, for the tDT construction, each full model is executed for only $N_{step}$ number of steps, where $N_{step}$ is usually a modest number. 

The choice of $N_{step}$ warrants further discussion. The requirement of $N_{step}\geq (n_M+2)$ is to ensure there are sufficient entries in the QoI time series \eqref{traj} to train the tDT FML model \eqref{tDT}. In many cases, $N_{step}$ often needs to be larger than that. This is because each full DT model simulation starts from a sampled initial condition \eqref{rand}, which is likely to be ``nonphysical" (especially when random sampling is employed). Therefore, it is necessary to allow the full DT model simulations to run for a period of time such that the solution states, as well as the QoIs, become "physical". Exactly how long the full DT model simulations need to be run is problem dependent. It is also worth noting that $N_{step}$ should not be excessively large, as the solutions may settle down to certain stationary states in such a way that the QoI time series \eqref{traj} contains little useful dynamical information.

\subsubsection{FML tDT Model Construction}

The FML tDT model takes the form \eqref{tDT}. It defines a mapping
$$
G:\mathbb{R}^{(n_M+1)\times n_V + n_\gamma} \to \mathbb{R}^{n_V},
$$ 
which maps $(n_M+1)$ consecutive entries in the QoI time series \eqref{traj}, along with the explicit parameters, to the $(n_M+2)$-th entry. To learn such the model, we seek a parameterized form
\be \label{G-theta}
V_{n_M+1}= G\left(V_{n_M},\dots, V_0; \gamma,\Theta\right),
\ee
where $\Theta$ are the hyper-parameters that need to be trained. To train this model, we need the QoI training time series data to have at least $(n_M+2)$ consecutive entries. Hereafter, such a QoI time series shall be called a QoI ``burst", whose length (i.e., number of consecutive entries) is
\be \label{burst}
n_L = (n_M+1) + n_R, \qquad n_R\geq 1,
\ee
where $n_R$ is the number of recurrent loss, or multi-step loss. To discuss the multi-step loss, let us first consider the standard one-step loss for training the FML tDT model \eqref{G-theta},
\be \label{loss1}
\mathcal{L}_1(\Theta) = \frac{1}{N_{data}}\sum_{j=1}^{N_{data}} \left\| V^{(j)}_{n_M+1} - G\left(V^{(j)}_{n_M},\dots, V^{(j)}_0; \gamma^{(j)},\Theta\right)\right\|^2,
\ee
where $N_{data}$ is the total number of the training data. Obviously, this is simply the mean-squared mismatch between the one-step FML model predictions and the training data.
For multi-step loss, let us consider the time marching scheme defined by the (untrained) FML model \eqref{G-theta},
\be
\left\{
\begin{aligned}
   & \widetilde{V}_{n_M+k} = G(\widetilde{V}_{n_M+k-1}, \dots, \widetilde{V}_{k-1}; \gamma, \Theta), \qquad k=1,\dots, \\
   & \widetilde{V}_0 = V_0, \quad\cdots, \quad\widetilde{V}_{n_M} = V_{n_M}, \quad\textrm{(Initial conditions)}
\end{aligned}
\right.
\ee
where $k\geq 1$ is the forward marching step. Let 
\be \label{loss-k}
\mathcal{L}_k(\Theta) = \frac{1}{N_{data}}\sum_{j=1}^{N_{data}} \left\| V^{(j)}_{n_M+k} - \widetilde{V}_{n_M+k}^{(j)}\right\|^2, \qquad k=1,\dots,
\ee
be the mean squared mismatch at the $k$-th step between the FML model prediction and the trajectory data \eqref{traj}. We then define the multi-step loss as
\be \label{loss}
\mathcal{L}(\Theta) \triangleq \frac{1}{n_R}\sum_{k=1}^{n_R} \mathcal{L}_k(\Theta),
\ee
where $n_R\geq 1$ is the total number of recurrent forward steps. It is obvious that the one-step loss $\mathcal{L}_1(\Theta)$ \eqref{loss1} is the special case of $n_R=1$. In practice, multi-step loss with $n_R = 5\sim 10$ can notably enhance the long-term numerical stability of FML models (\cite{ChurchillXiu_FML2023}).

Upon taking into consideration of the multi-step loss \eqref{loss}, each piece of the training data is a time series QoI burst of length $n_L$ \eqref{burst}. They are to be selected from the QoI dataset $\mathcal{Q}$ \eqref{traj}, where each QoI trajectory has a length $N_{step}$. In most cases, we have $N_{step}> n_L$, and often $N_{step}\gg n_L$. It is therefore possible, and preferred, to choose multiple QoI bursts out of each QoI trajectory from $\mathcal{Q}$. Let $n_B\geq 1$ be the number of QoI bursts data chosen from each trajectory of $\mathcal{Q}$. We propose to use random sampling with uniform distribution to choose the $n_L$-length burst data out of the $N_{step}$-length QoI trajectory, resulting in a total number of $n_B\cdot N_{sim}$ of QoI burst data. These become the tDT training dataset:
\be \label{dataset}
\mathcal{D} \triangleq \left\{\left. V_0^{(j)}, \dots, V_{n_M}^{(j)}, \dots, V_{n_M+n_R}^{(j)}; \gamma^{(j)}\right|j=1,\dots, N_{data}\right\},
\ee
where
\be \label{n-data}
N_{data} = n_B \cdot N_{sim}
\ee
is the total number of training data. Since each trajectory data in $\mathcal{Q}$ \eqref{traj} follows the ``same" dynamical setting, we advocate not to use a large number of $n_B$. In practice, $n_B \sim 10$, or less, is preferred. More discussion of this can be found in \cite{ChurchillXiu_FML2023}.

\begin{figure}[htbp]
	\begin{center}
        \includegraphics[width=0.8\textwidth]{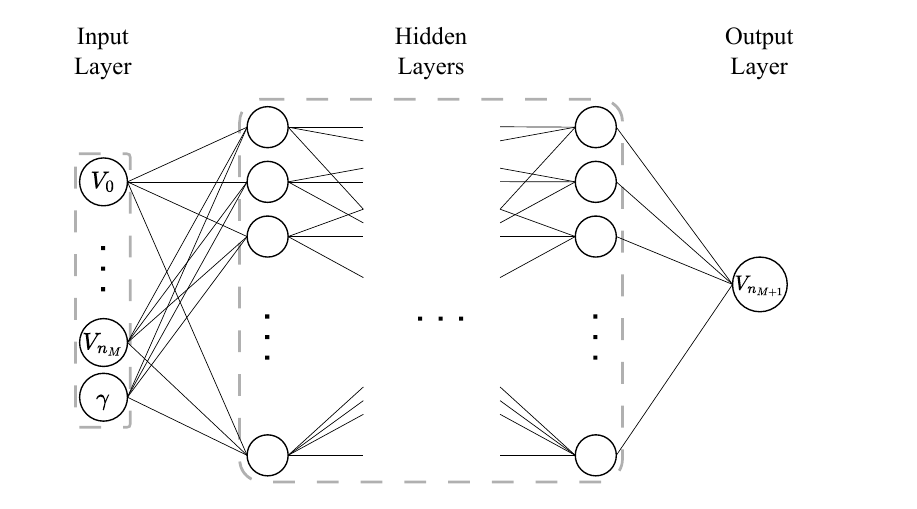}
		\caption{DNN structure for FML targeted Digital Twin model.}
\label{fig:DNN}
	\end{center}
\end{figure}

Minimization of the loss function \eqref{loss} over the training data set \eqref{dataset} results in fixed values for the hyper-parameters $\Theta$ in the FML model \eqref{G-theta}. We then obtain a trained FML tDT model
\be \label{G-trained}
V_{n_M+1}= G\left(V_{n_M},\dots, V_0; \gamma\right),
\ee
where the fixed hyper-parameter $\Theta$ is suppressed.
Figure \ref{fig:DNN} illustrates the network structure when DNN is used to construct the FML model. Note that in most cases, a simple feedforward fully connected network is sufficient.

\subsection{System Analysis via tDT}

The trained tDT model \eqref{G-trained} provides a prediction and analysis tool for the dynamics of the QoI, without requiring simulations of the full DT model. More specifically, the predictive QoI model by the tDT can be written as
\be \label{tDT-model}
\left\{
\begin{aligned}
   & V_{n+1}= G\left(V_{n},\dots, V_{n-n_M}; \gamma\right), \quad n\geq n_M, \\
   & \textrm{Initial conditions:} \quad V_0, \cdots, V_{n_M}. 
\end{aligned}
\right.
\ee

Given a time series of the QoIs, $V_0, \cdots, V_{n_M}$, as ``initial conditions", the FML tDT model can be marched forward in time, like a multi-step time integrator, to predict the future states of the QoIs. The ``initial conditions" can be acquired from either the full DT simulations or data collected from the physical twin. Once the initial conditions are prescribed, the tDT is synchronized with the full DT, in the sense that the time series $\left(V_0, \cdots, V_{n_M}\right)$ should correspond to certain unknown values of the hidden parameters $\hat{\gamma}$ \eqref{gamma}. The tDT would conduct its prediction of QoI dynamics under the same unknown value of the hidden parameters.

\section{Application to Fluid Dynamics} \label{sec:examples}

We now apply the proposed tDT framework to a computational fluid dynamics (CFD) problem.
In particular, we consider the well-studied problem of two-dimensional incompressible flow past a circular cylinder.
Although not excessively complicated, the problem possesses sufficient computational complexity to demonstrate the effectiveness of the proposed tDT approach.

\subsection{The Full Digital Twin}

\begin{figure}[htbp]
 	\begin{center}
 		\includegraphics[width=0.45\textwidth]{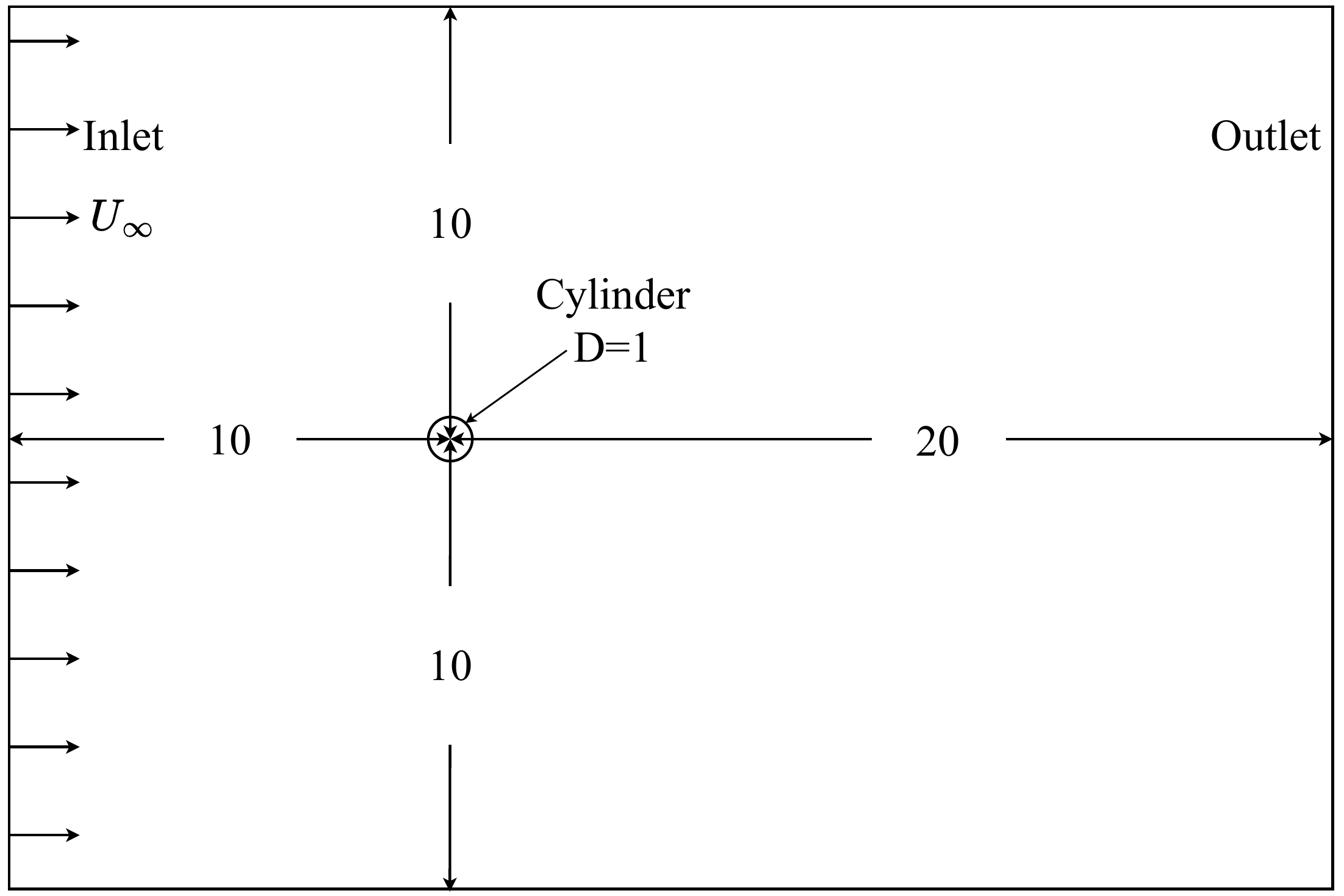}
        \includegraphics[width=0.45\textwidth]{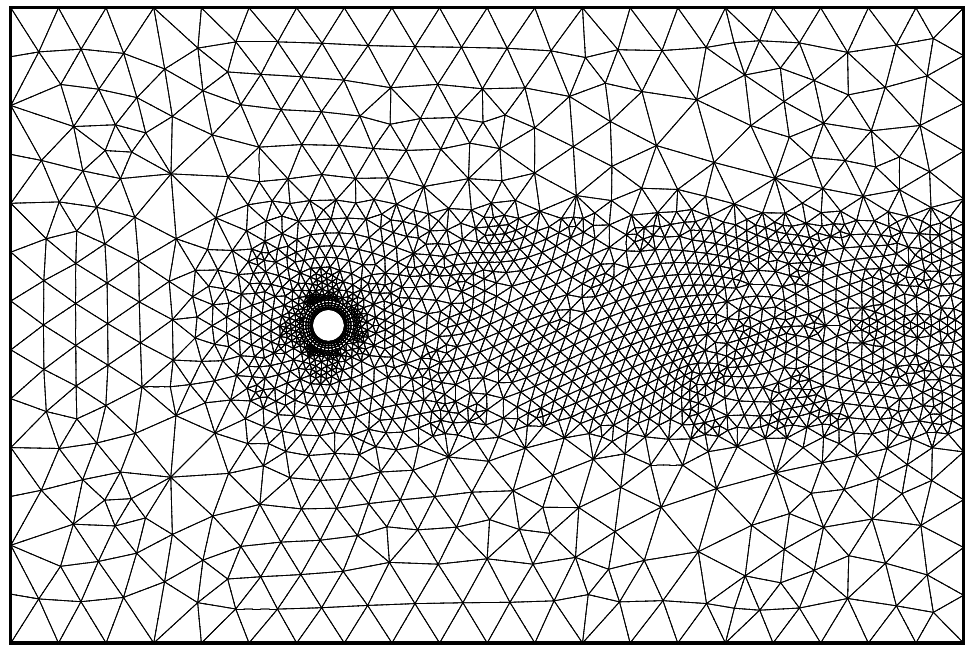}
 		\caption{Setup for the flow past cylinder simulation. Left: computational domain; Right: computational mesh.}
 		\label{fig:diagram}
 	\end{center}
 \end{figure}

The governing equations are the non-dimensional incompressible Navier-Stokes equations for velocity $\mathbf{u} = (u,v)^T$ and pressure $p$:
\be
\left\{
  \begin{aligned}
   & \nabla\cdot\mathbf{u} = 0, \\
&\mathbf{u}_t + (\mathbf{u}\cdot \nabla)\mathbf{u} = -\nabla p +
\frac{1}{Re}\nabla^2\mathbf{u},
\end{aligned}
\right.
\ee
where $Re$ is the Reynolds number
\be
Re = \frac{U_\infty L}{\nu},
\ee
with $U_\infty$ being characteristic velocity, $L$ characteristic length, and $\nu$ the kinematic viscosity. The domain is a rectangular region $(x,y)\in [-10,20]\times [-10,10]$, where a cylinder with diameter $D=L=1$ is centered at the origin. For boundary conditions, far-field conditions of $u = U_\infty = 1$ and $v=0$ are imposed on the left (inlet), top, and bottom boundaries, while an outflow condition of ($\partial_x = 0$) is enforced on the right boundary (outlet).
The computational domain is illustrated on the left of Figure~\ref{fig:diagram}.

To solve the problem numerically, we employ Nektar++  (\cite{cantwell2015nektar++,moxey2020nektar++}), an open-source high-order $hp$-spectral element solver. The domain is discretized by an unstructured mesh of $3,613$ elements, shown on the right of Figure~\ref{fig:diagram}. Spectral elements of 6th-order polynomial are used, yielding a total number of $143,880$ degrees of freedom for the discrete system. That is, in the full DT system \eqref{DT}, $N_U=143,880$. Second-order implicit–explicit scheme is used for time integration, with a constant step size $\delta t=0.002$. All of these parameter settings were examined numerically to ensure sufficient resolution of the problem.

%

\subsection{The Quantity-of-Interest}

For flow past a body or a structure, one is often interested in the force exerted on the structure by the flow. Therefore, we consider two separate cases for the QoIs:
(i) the total drag and lift forces on the cylinder, and (ii) the pressure distribution on the surface of the cylinder. Hereafter, we shall construct two tDT models corresponding to the two cases.

\subsubsection{Case 1: Drag and Lift Forces} \label{sec:QoI-1}

In the first case, we consider the overall force acting on the cylinder 
$$
    \mathbf{f} = (f_x, f_y)^T = - \oint_S \left(\boldsymbol{\sigma}\cdot\hat{\mathbf{n}}\right) ds,
$$
where the integral is over the cylinder surface $S$,
$$
    \boldsymbol{\sigma} = -p \mathbf{I} + \nu\left( \nabla \mathbf{u} + \nabla \mathbf{u}^{T} \right)
$$
the Cauchy stress tensor, and
$\hat{\mathbf{n}}$ the outward unit normal vector. In our coordinate setting, $f_x$ becomes the drag force and $f_y$ the lift force.

It is a standard practice to normalize the drag and lift, resulting in the drag and lift coefficients.
$$
    C_D = \frac{f_x}{\frac{1}{2} U_\infty^2 L}, 
    \qquad 
    C_L = \frac{f_y}{\frac{1}{2}U_\infty^2 L}.
$$
These become the QoIs in the full DT system \eqref{fullDT},
\be \label{V1}
    V = (C_D,\, C_L)^T = (2f_x, 2f_y)^T\in\mathbb{R}^2,
\ee
where the factor of 2 is the result of our chosen characteristic velocity and length.

\subsubsection{Case 2: Surface Pressure on Cylinder} \label{sec:QoI-2}

In the second case, we consider the pressure distribution along the surface $S$ of the cylinder as the QoI. Since the surface is a circle, we parameterize the pressure by a truncated Fourier series,
\be \label{Ps}
P_S(\theta)=a_0+\sum_{n=1}^N a_n \cos (n \theta)+b_n \sin (n \theta),
\ee
where $\theta = \arctan (y/x)$ is the polar angle along the surface. Upon numerical experimentation, we determined that $N=30$ is sufficient to fully resolve the pressure distribution function on the surface. The Fourier expansion coefficients become the QoIs in this case, i.e.,
\begin{equation} \label{V2}
V = \big( a_0,\, a_1,\dots,a_N,\, b_1,\dots,b_N \big) \in \mathbb{R}^{2N+1}.
\end{equation}

\subsection{Trajectory Data Generation}

Once the QoIs are determined in the two separate cases, \eqref{V1} and \eqref{V2}, we then seek to generate training datasets, in the form of \eqref{dataset}, for the two cases.

For the full DT \eqref{fullDT}, the only system parameter is the Reynolds number, i.e., $p = \{Re\}$ and $q=\emptyset$. In this paper, we consider the $Re$ number range to be 
\be \label{Re}
Re \in I_P = (100, 2000).
\ee
We then conduct $N_{sim} = 7,800$ full DT simulations, each of which uses a uniformly distributed random $Re$ number in the range \eqref{Re} and is conducted for over $0 \leq t \leq 200$ with ``sudden start" initial conditions of $u_0=1$ and $v_0=0$. 

We choose the explicit parameter set \eqref{gamma} to be empty, $\gamma = \emptyset$. This implies that the $Re$ number is treated as a hidden parameter in the tDT models and not recorded during the full DT simulations.
The QoIs for the two cases are then recorded during the simulations over a constant time step $\Delta t= 0.1$, which is determined to be sufficient to resolve the temporal scale of the QoIs. We then obtain the trajectory dataset \eqref{traj} with $N_{step}=2,000$, i.e.,
\be \label{cylin_traj}
\mathcal{Q} = \left\{\left.V_0^{(j)}, V_1^{(j)}, \dots, V_{2000}^{(j)}\right\vert j=1,\dots, 7800\right\}.
\ee
Again, we emphasize that the $Re$ number is not recorded in these trajectory data. By doing so, the trained FML tDT model is capable of conducting QoI predictions without ``knowing" the $Re$ number. This choice of ``hiding" the $Re$ number is to mimic the more practical setting, where the precise value of the $Re$ is often difficult to determine. 
For FML, it is perfectly fine to keep the $Re$ number as an explicit input. This will introduce an additional node in the input layer of the DNN structure. The training data will need to retain the $Re$ number of each trajectory. The trained FML model will then conduct QoI predictions when a $Re$ number and the initial conditions are prescribed. 
More details of FML modeling with hidden parameters can be found in \cite{FuEtal_JMLMC22}.

\subsection{Training Data}

Upon extensive numerical experimentation, we determined that a memory term $n_M=49$ is sufficient to construct accurate tDT models for both cases. Also, the number of multi-step loss \eqref{loss} with $n_R=10$ was deemed to be sufficient. This results in a requirement of the training data to have a length of $n_L = 60$ \eqref{burst}. We then conduct uniform random sampling of $n_L$-length burst data from the trajectory dataset \eqref{cylin_traj}. From each of the $N_{sim}=7,800$ trajectories, we sample $n_B$ segments of $n_L$-length data and obtain a total number of $N_{data}=n_B N_{sim}$ pieces of training data. 

To summarize, the training datasets for both Case 1 and Case 2 take the following form
\be \label{cylin_data}
\mathcal{D} = \left\{\left. V_0^{(j)}, \dots, V_{n_M}^{(j)}, \dots, V_{n_M+n_R}^{(j)}\right|j=1,\dots, N_{data}\right\},
\ee
where $n_M=49$ and $n_R=10$ for both cases, and for Case 1, $n_B=10$ and $N_{data} = 78,000$; for Case 2 $n_B=20$ and $N_{data}= 156,000$. With $\Delta t=0.1$, each piece of the training data spans a time interval of $t=n_L\Delta t =6.0$, with the memory term spanning $(n_M+1)\Delta t = 5.0$ and the multi-step loss spanning $n_R\Delta t = 1.0$.

\begin{figure}[htbp]
	\begin{center}

        \includegraphics[width=\textwidth]{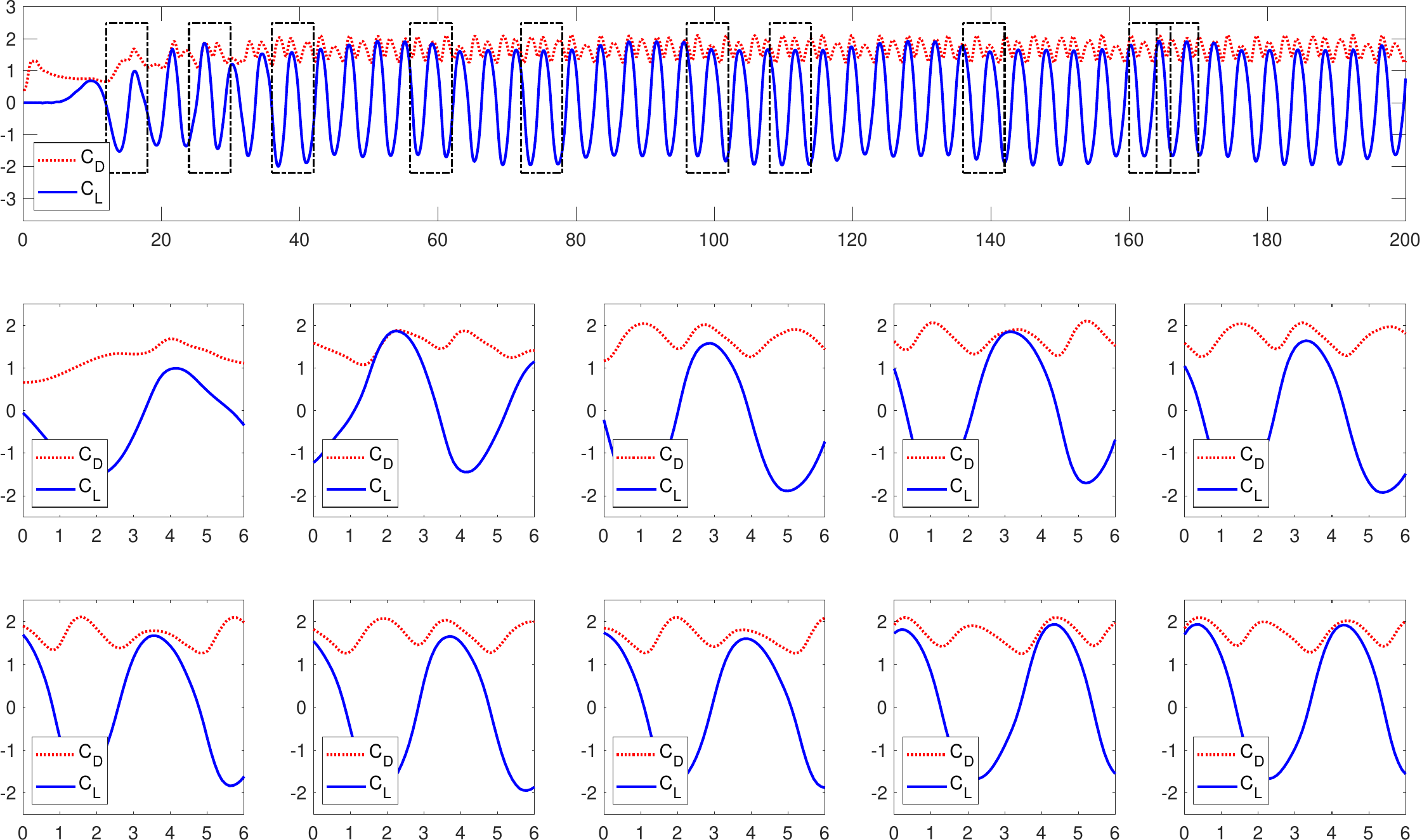}
		\caption{Illustration of $10$ training data sampled from a single full DT simulation trajectory for $t\in[0, 200]$ at $Re=1997.055$. Note that the $Re$ number is not recorded.}
\label{fig:data1}
	\end{center}
\end{figure}

In Figure~\ref{fig:data1}, we illustrate the procedure for constructing the training data for Case 1, where the drag and lift coefficients $C_D$ and $C_L$ \eqref{V1} are the QoIs. The top figure in Figure~\ref{fig:data1} shows one trajectory data of $(C_D, C_L)$ from the full DT simulation in the time domain $t\in [0,200]$. The Reynolds number $Re=1997.055$ is randomly sampled in the range \eqref{Re} but not recorded. The 10 rectangular boxes in the figure depict the $n_B=10$ randomly sampled segments of length $t=6.0$. The ``starting" and ``ending" times of these segments are not recorded. Instead, each segment is recorded in the time domain $0\leq t\leq 6.0$. These segments, plotted in the bottom two rows of Figure~\ref{fig:data1}, become 10 pieces of training data in the final training data set for Case 1.

\begin{figure}[htbp]
	\begin{center}

        \includegraphics[width=\textwidth]{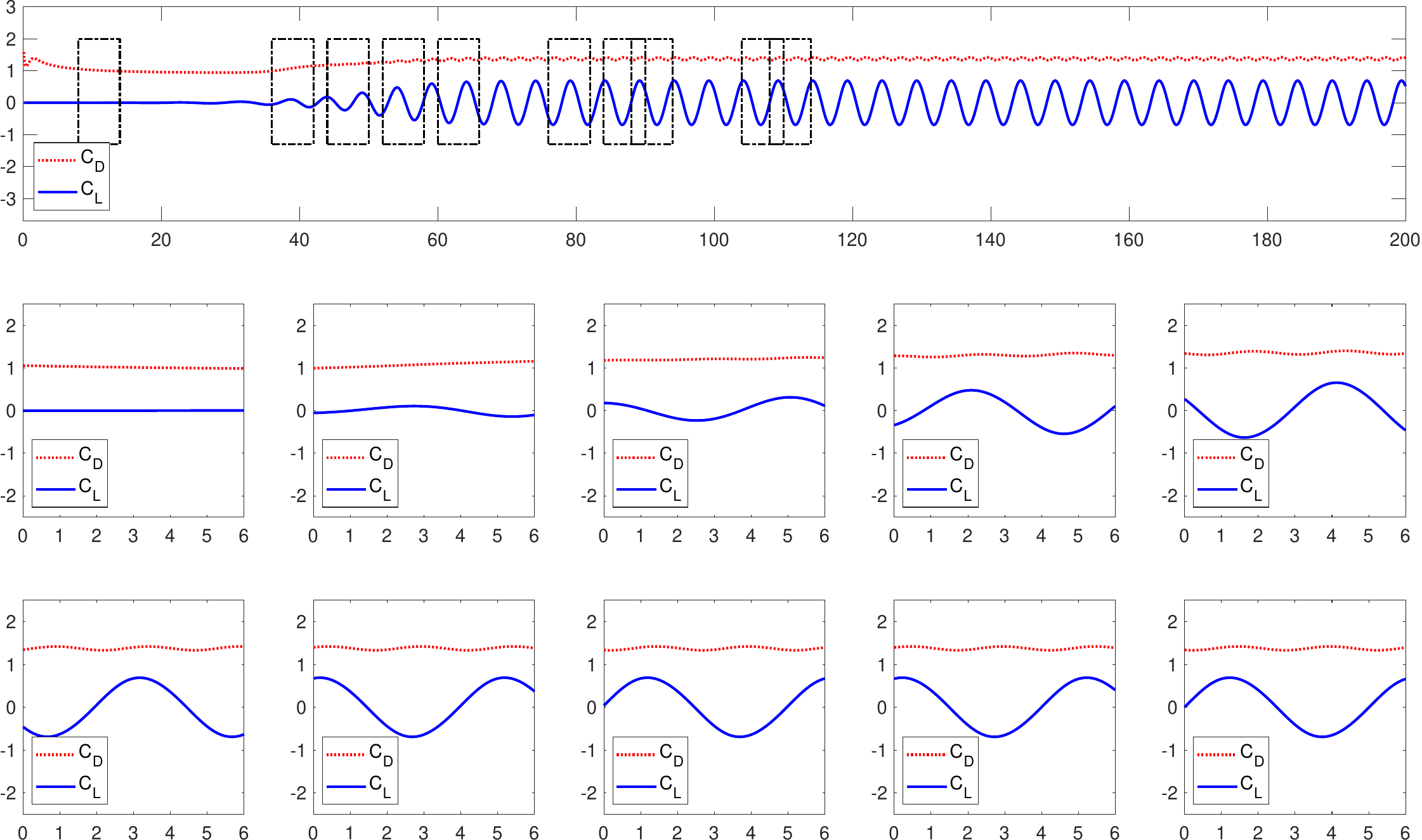}
		\caption{Illustration of $10$ training data sampled from a single full DT simulation trajectory for $t\in[0, 200]$ at $Re=195.763$. Note that the $Re$ number is not recorded.}
\label{fig:data2}
	\end{center}
\end{figure}

Figure~\ref{fig:data2} demonstrates the same procedure for extracting another 10 segments of training data for Case 1 from another full DT simulation at Reynolds number $Re=195.763$, which is randomly sampled in the range \eqref{Re} but not recorded. 
This procedure is repeated for each of the $N_{sim}=7,800$ full DT simulations and results in the $N_{data}=78,000$ segments of training data \eqref{cylin_data} for Case 1.

For Case 2, the procedure for constructing the training data set \eqref{cylin_data} is the same, with only two differences compared to Case 1. The first difference is that the trajectory data \eqref{cylin_traj} contain the $(2N+1=61)$ Fourier coefficients of the surface pressure \eqref{V2} as the QoIs, and the second difference is that we sample $n_B=20$ random segments from each trajectory data to result in a total number of $N_{data}=156,000$ segments of training data.
   
\subsection{FML tDT Models}

The FML tDT models for both Case 1 and Case 2 are fully connected feedforward DNNs. See Figure~\ref{fig:DNN} for the network structure, where $n_M=49$. Note that since there is no explicit parameter, the input nodes for $\gamma$ are not required. 

For Case 1, the input layer contains the nodes representing $V=(C_D, C_L)^T$ from $V_0$ to $V_{n_M}$, resulting in a total of 100 nodes. The output layer contains 2 nodes, one for $C_D$ and the other for $C_L$. In between, we employ 5 hidden layers, each of which consists of 50 nodes. In summary, the DNN tDT model for Case 1 has layers with width $\{100, 50, 50, 50, 50, 50, 2\}$. 

For Case 2, the QoIs $V$ are the $61$ Fourier coefficients \eqref{V2}. The input layers contain these QoIs from $V_0$ to $V_{n_M}$, resulting in 3,050 nodes. The output layer contains 61 nodes for the QoIs. In between, we employ 5 hidden layers with width $\{256,64,64,64,64\}$. In summary, the DNN tDT model for Case 2 has layers with width $\{3050, 256, 64, 64, 64, 64, 61\}$. Numerical experimentation was conducted to verify that these chosen structures are sufficient to deliver satisfactory numerical results.

The DNN training was performed using the Adam optimizer to minimize the multi–step loss \eqref{loss} over $n_R=10$ steps, with batch size $64$ for $3,000$ epochs. A cyclic learning rate schedule was employed, with a base rate $10^{-7}$, maximum rate $10^{-3}$, and decay factor $0.999997$.

Hereafter, we shall call the trained tDT model for Case 1 ``tDT-1", and the trained tDT model for Case 2 ``tDT-2".

\subsection{Validation and Prediction of the tDTs}

We now present numerical tests to demonstrate the performance of the trained tDTs. Both tDT models take the form of \eqref{tDT-model}, which requires $(n_M+1=50)$ steps of initial conditions for a time history of the QoIs spanning $t=5.0$. The initial conditions are obtained from a full DT simulation at a specific $Re$ number and during a specific physical time window of $[\hat{t}, \hat{t}+5.0]$. Hereafter, we shall use $\hat{t}$ to denote the ``physical" time used in the full DT simulations. Once the initial conditions are loaded into the tDT models, they become ``synchronized" with the full DT, in the sense that the tDT models will predict the dynamics of the QoIs starting from the specific physical time $(\hat{t}+5.0)$ and at the specific $Re$ number, even though the $Re$ number and the physical time $\hat{t}$ are never recorded and stored in the training data and are not in the inputs for the tDT models. 

For the examples presented in this section, the $Re$ numbers are randomly sampled from the range \eqref{Re} and the starting physical time $\hat{t}$ is also randomly sampled from $[0,200]$. The tDT model predictions are conducted for long-term, for up to at least $0\leq t\leq 200.0$. For clarity of presentation, we shall plot the tDT predictions in terms of the physical time $\hat{t}$. Again, we emphasize that the tDT models are ``unaware" of the existence of the physical time $\hat{t}$.

\subsubsection{Case 1: tDT-1 for Drag and Lift Coefficients}

\begin{figure}[htbp]
	\begin{center}
        \includegraphics[width=0.4\textwidth]{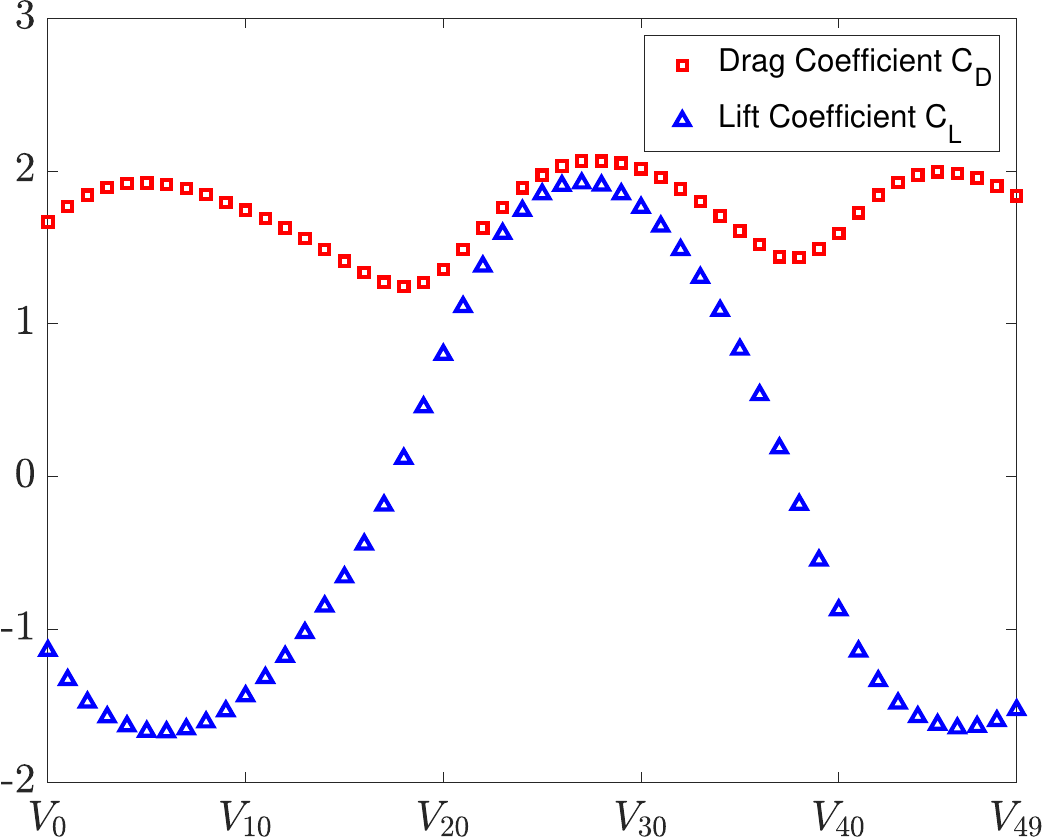}
		\caption{Test 1-1: The initial conditions for the tDT-1 model. (Generated by the full DT with $Re=1,900$ and during physical time $195.0\leq \hat{t}\leq 200.0$.)}
\label{fig:IC1}
	\end{center}
\end{figure}

In the first test case (Test 1-1), we employ initial conditions for the drag and lift coefficients computed from the full DT simulation at $Re=1,900$ and during $195.0\leq \hat{t}\leq 200.0$.
The initial conditions are shown in Figure~\ref{fig:IC1}, where the horizontal axis demonstrates how the initial conditions are loaded into the tDT-1 model \eqref{tDT-model}.
Once the initial time history is loaded into the tDT-1 model for $0\leq t\leq 5.0$,  it starts the prediction of $(C_D, C_L)$ from $t=5.0$, which corresponds to the (unknown) physical time $\hat{t}=200.0$.
Note that the $Re=1,900$ is specifically chosen to be not within the training dataset \eqref{cylin_data}. 

\begin{figure}[htbp]
	\begin{center}
        \includegraphics[width=\textwidth]{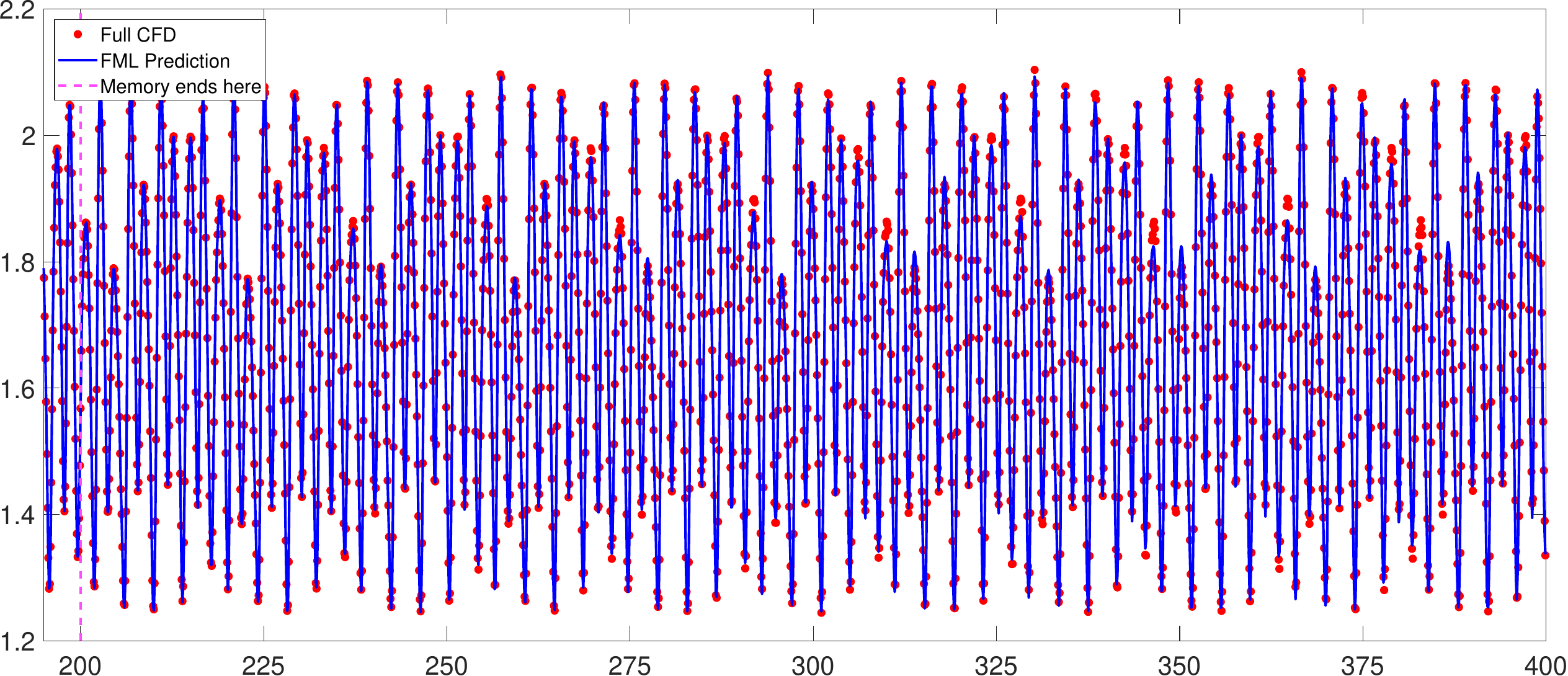}
        \includegraphics[width=\textwidth]{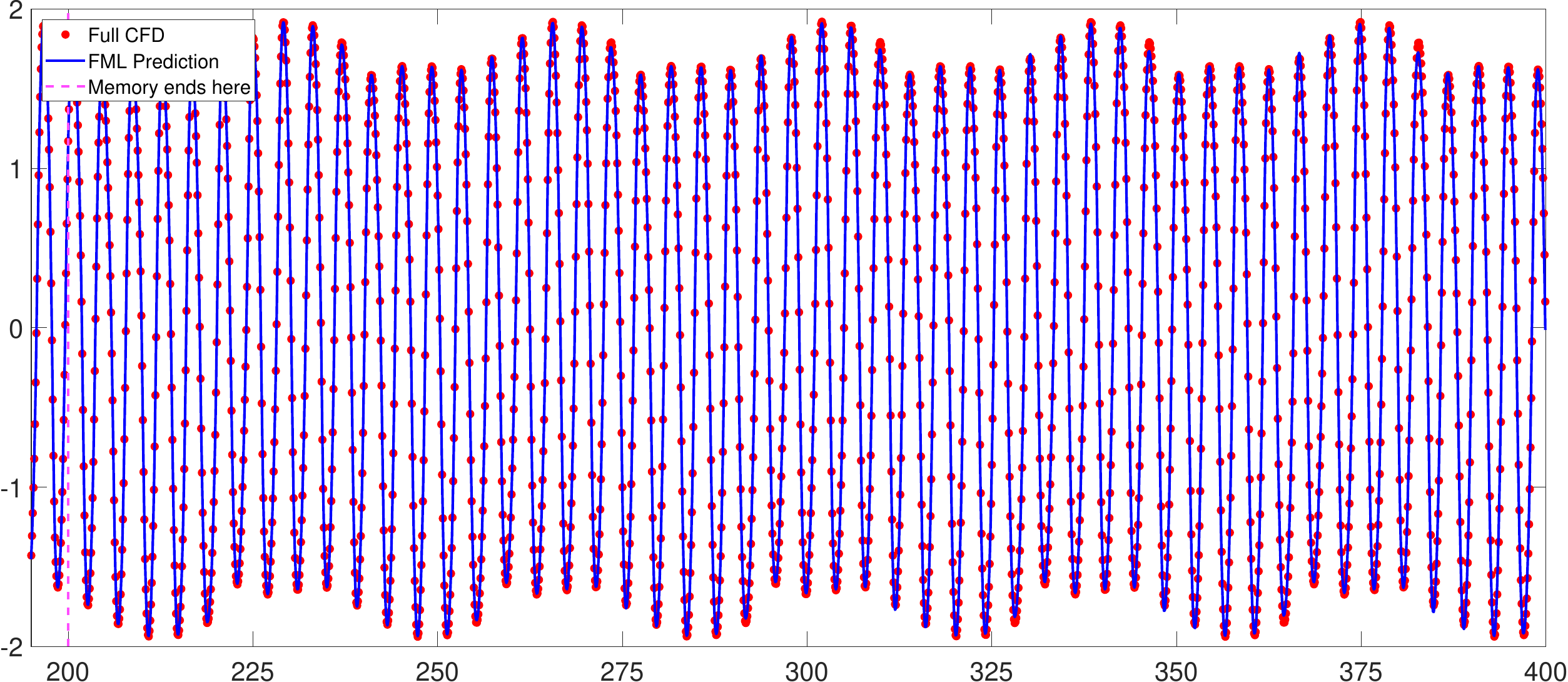}
		\caption{Test 1-1: Prediction by tDT-1 for up to $t=200.0$, which corresponds to physical time up to $\hat{t}=400.0$, in comparison with the reference solution by the full CFD DT simulation. Top: Drag coefficient $C_D$; Bottom: Lift coefficient $C_L$.}
\label{fig:tDT-1-1}
	\end{center}
\end{figure}

\begin{figure}[htbp]
	\begin{center}
        \includegraphics[width=\textwidth]{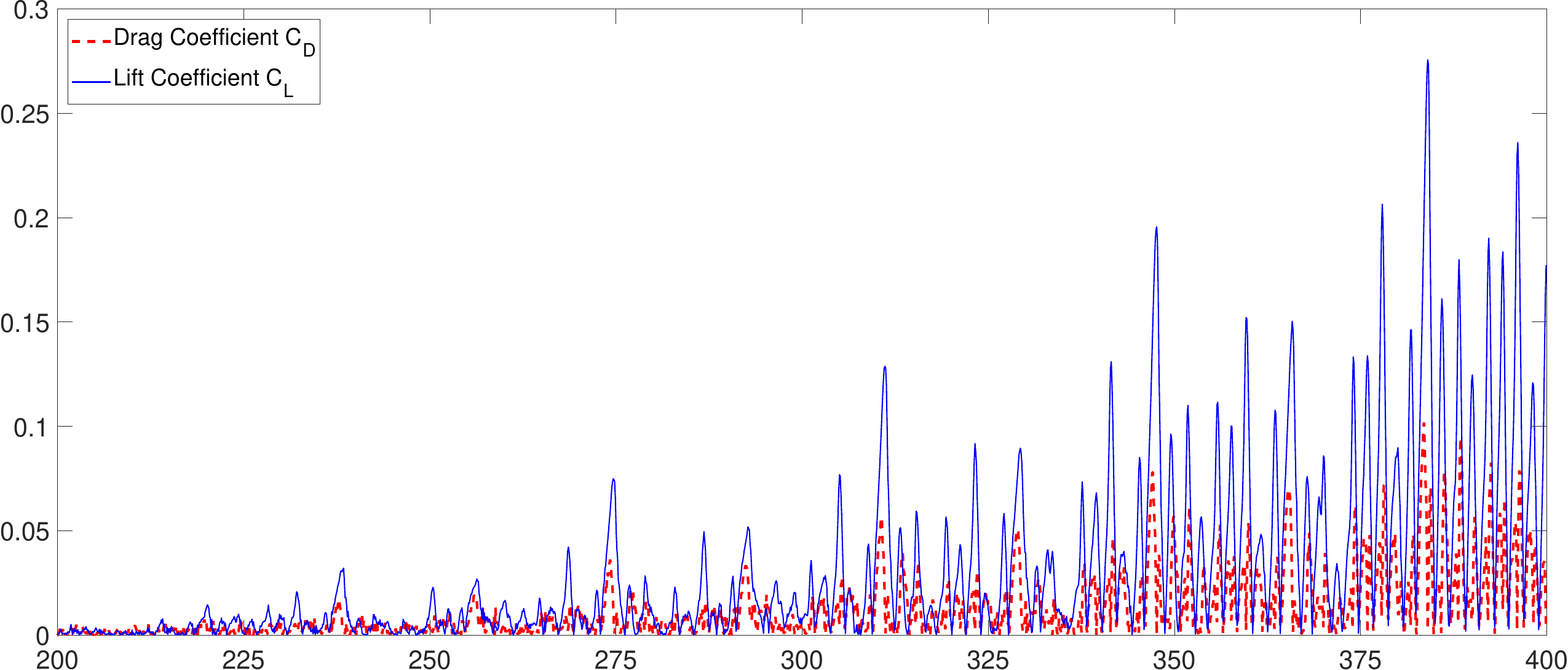}
		\caption{Test 1-1: Errors of the tDT-1 model prediction of $C_D$ and $C_L$ against the reference full CFD solution.}
\label{fig:err-1-1}
	\end{center}
\end{figure}

\begin{figure}[htbp]\centering
\begin{subfigure}{0.58\textwidth}
    \begin{center}
        \includegraphics[width=1\textwidth]{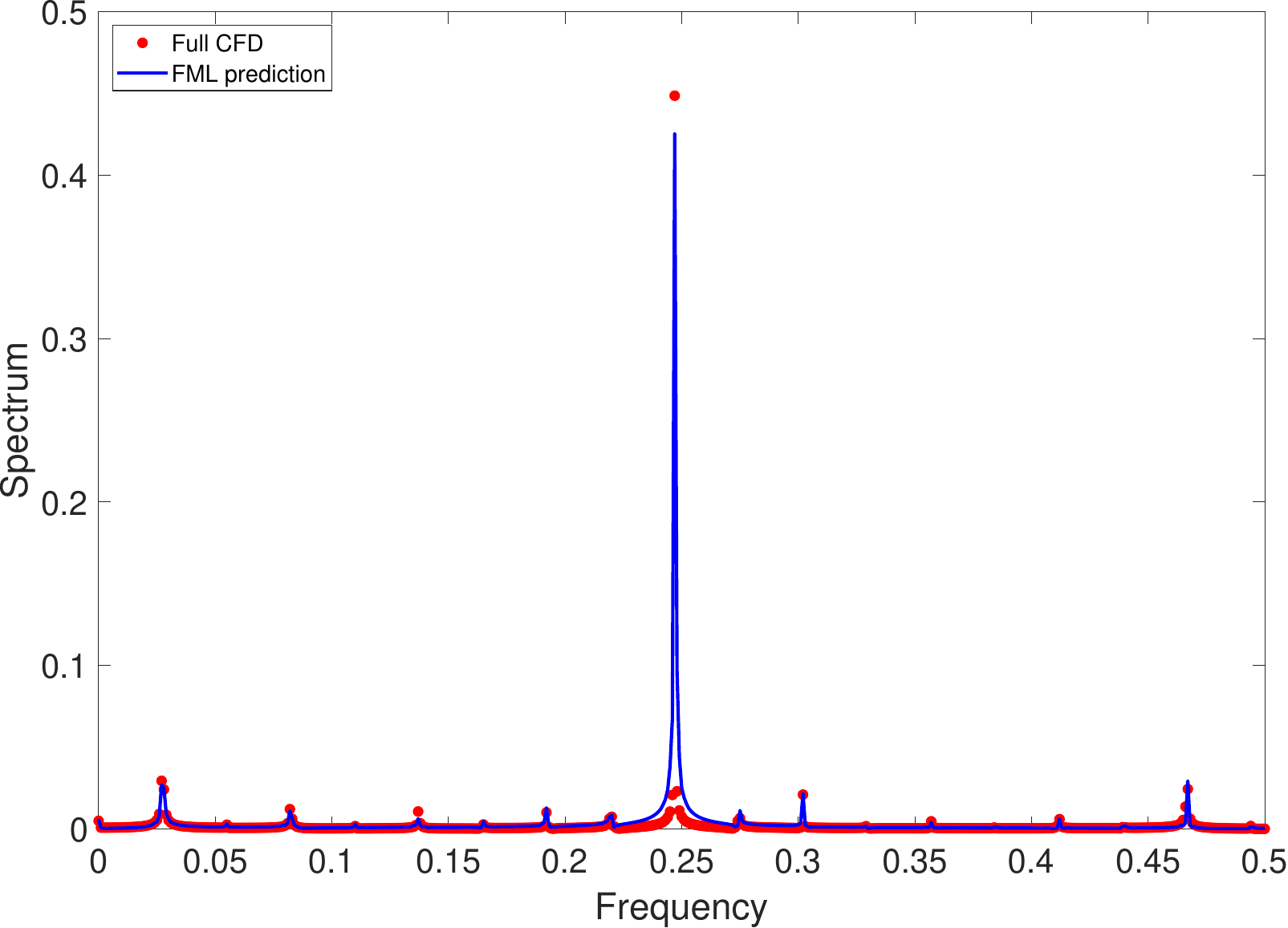}
\label{fig:spec}
	\end{center}
\end{subfigure}\\
	\begin{subfigure}[t]{0.9\textwidth}
    \begin{center}
\begin{tabular}{c|cccc}
\hline
Rank of frequency    & 1     & 2     & 3     & 4     \\\hline
Full CFD       & 0.247 & 0.028 & 0.467 & 0.302 \\\hline
FML Prediction & 0.247 & 0.027 & 0.467 & 0.302\\\hline
\end{tabular}
\label{tab:spec}\end{center}
    \end{subfigure}
    \caption{Test 1-1: Comparison of the spectra of $C_L$ signal between the tDT-1 model and the full CFD DT model.}
    \label{fig:spec-1-1}
\end{figure}

The prediction of the tDT-1 model is performed for a long-term simulation of up to $t=200.0$, which corresponds to the physical time of up to $\hat{t}=400.0$. The predictions of $C_D$ and $C_L$ are shown in Figure~\ref{fig:tDT-1-1}, where the solid line is the prediction by the tDT-1 model and the circles are the reference solution obtained by the full DT simulation. We observe very good agreement between the tDT-1 predictions and the full DT predictions. The errors of the tDT predictions are shown in Figure~\ref{fig:err-1-1}, where we observe a slow increase in the errors. This is to be expected, as for limit cycle type of solutions in this case, a small phase error (which is unavoidable by any numerical methods) shall always induce a growth in point-wise error over time. To examine the phase error, we conduct spectrum analysis of the $C_L$ signal and report the results in Figure~\ref{fig:spec-1-1}. The dominating frequencies $\hat{f}$ correspond to Strouhal numbers $St=\hat{f}L/U_\infty$. We observe excellent agreement between the tDT-1 model prediction and the reference solution by the full DT simulation. For the first four dominating frequencies, the tDT-1 model predictions match the reference values for up to 3 significant digits. This demonstrates the excellent accuracy of the tDT prediction. It is worth noting that in this case, the simulation time of the tDT-1 model prediction is on the order of seconds, whereas the full DT simulation is on the order of hours.

\begin{figure}[htbp]
	\begin{center}
        \includegraphics[width=0.4\textwidth]{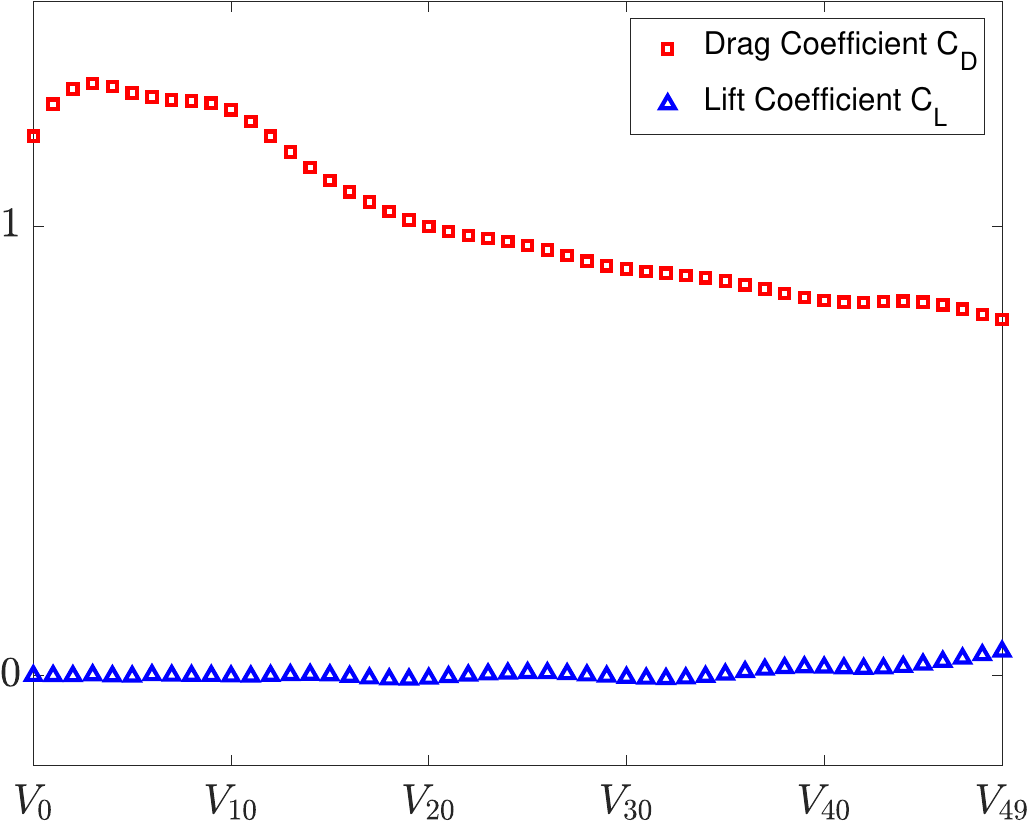}
		\caption{Test 1-2: The initial conditions for the tDT-1 model. (Generated by the full DT with $Re=1,750$ and during physical time $1.0\leq \hat{t}\leq 6.0$.)}
\label{fig:IC2}
	\end{center}
\end{figure}

\begin{figure}[htbp]
	\begin{center}
        \includegraphics[width=\textwidth]{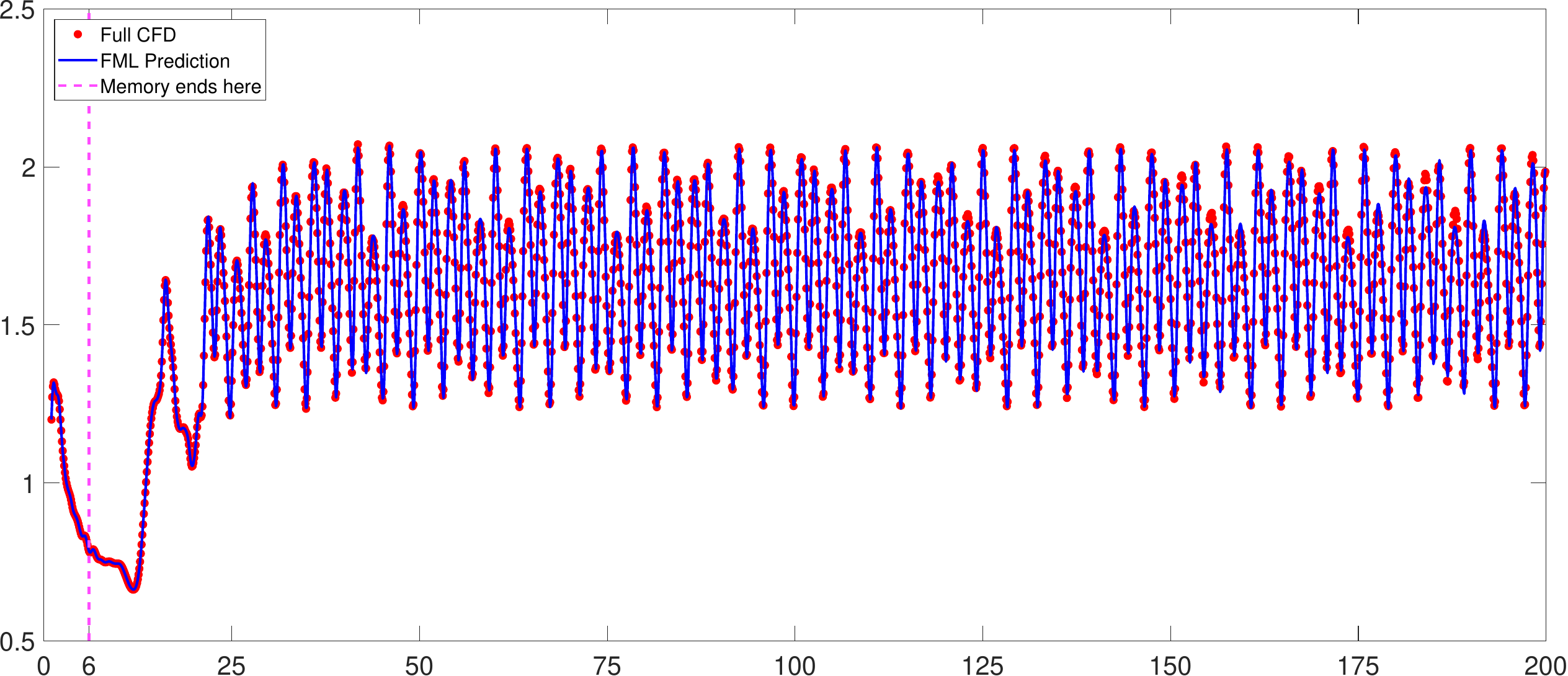}
        \includegraphics[width=\textwidth]{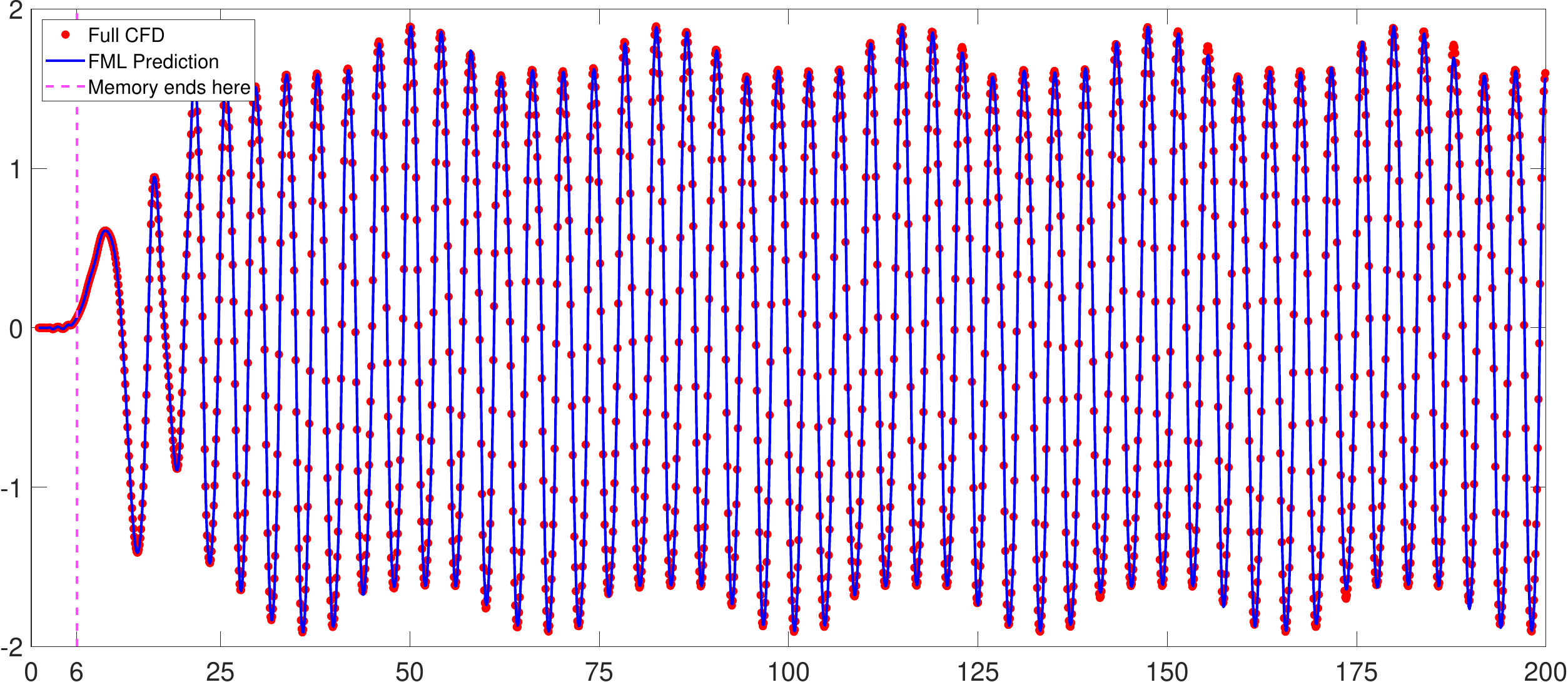}
		\caption{Test 1-2: Prediction by tDT-1 for up to $t=194.0$, which corresponds to physical time up to $\hat{t}=200.0$, in comparison with the reference solutions by the full CFD DT simulation. Top: Drag coefficient $C_D$; Bottom: Lift coefficient $C_L$.}
\label{fig:tDT-1-2}
	\end{center}
\end{figure}

\begin{figure}[htbp]
	\begin{center}
        \includegraphics[width=\textwidth]{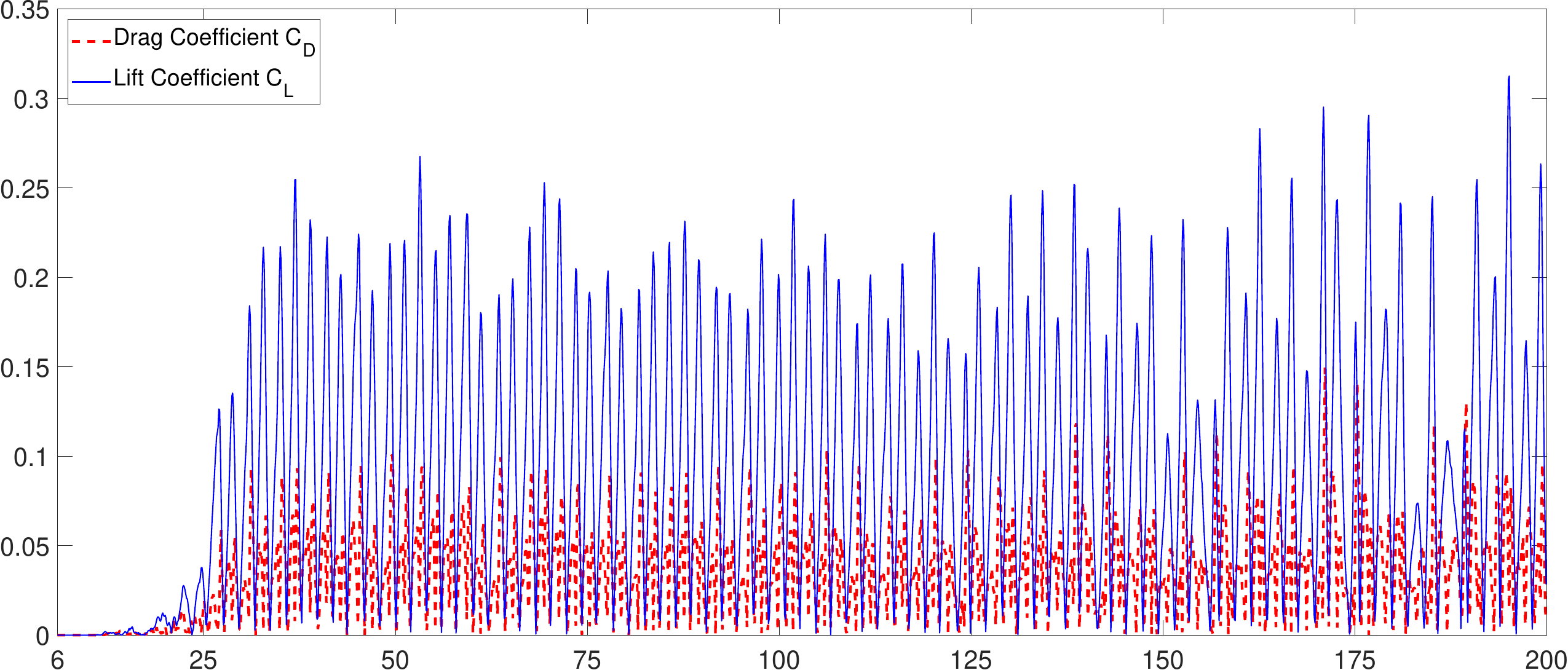}
		\caption{Test 1-2: Errors of the tDT-1 model prediction of $C_D$ and $C_L$ against the reference full CFD solution.}
\label{fig:err-1-2}
	\end{center}
\end{figure}

In the second test (Test 1-2), we employ an initial condition from a very early stage of the full DT simulation. In particular, we use the $(C_D, C_L)$ time history from physical time $1.0\leq \hat{t}\leq 6.0$ with $Re=1,750$ as the initial condition, shown in Figure~\ref{fig:IC2}. This corresponds to a flow shortly after the initial ``sudden start" condition, when the fluid flow will go through a transit period and far from settling down to quasi-periodic vortex shedding regime. None of the information about the ``physics" is available to the tDT-1 model, as it requires only the information for the initial conditions in  Figure~\ref{fig:IC2}. 
The tDT-1 model predictions for $0\leq t\leq 194.0$, which corresponds to the physical time $6.0\leq \hat{t}\leq 200.0$, are shown in Figure~\ref{fig:tDT-1-2}, where we again observe good agreements with the reference solution by the full DT simulation. The tDT-1 model, without requiring any information about the fluid field, is able to predict the dynamics of drag and lift forces from the early transition states to the later quasi-periodic states. The accuracy of the tDT-1 predictions is examined in Figure~\ref{fig:err-1-2}, where the point-wise errors remain stable throughout the prediction time horizon.

Many other test cases were examined at different $Re$ numbers and different initial conditions in physical time. The results are similar to the two cases presented here.

\subsubsection{Case 2: tDT-2 for Surface Pressure Distribution}

We now examine the performance of the tDT-2 model, whose QoIs are the Fourier expansion coefficients of the pressure distribution along the cylinder surface \eqref{V2}. When supplied with proper initial conditions, the tDT-2 model can predict the dynamics of the Fourier coefficients over time, which in turn provides predictions of the surface pressure distribution via the Fourier expansion \eqref{Ps}. 

In the first test case (Test 2-1), we choose the tDT-2 initial conditions from the full DT simulation with $Re=1,500$ and during physical time $195.0\leq \hat{t}\leq 200.0$. 
We then conduct the tDT-2 model prediction for $0.0\leq t\leq 285.0$, which corresponds to the physical time $200.0\leq \hat{t}\leq 485.0$.

\begin{figure}[htbp]
	\begin{center}     
\begin{subfigure}[t]{0.9\textwidth}
\includegraphics[width=\textwidth,]{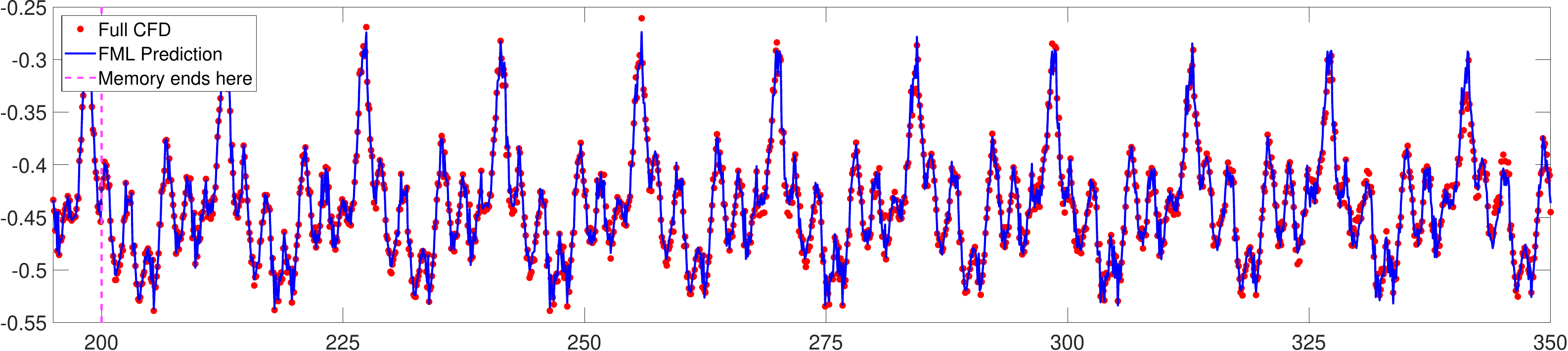} 
    \caption{Coefficient $a_0$}
    \end{subfigure}    
\begin{subfigure}[t]{0.9\textwidth}
\includegraphics[width=\textwidth,]{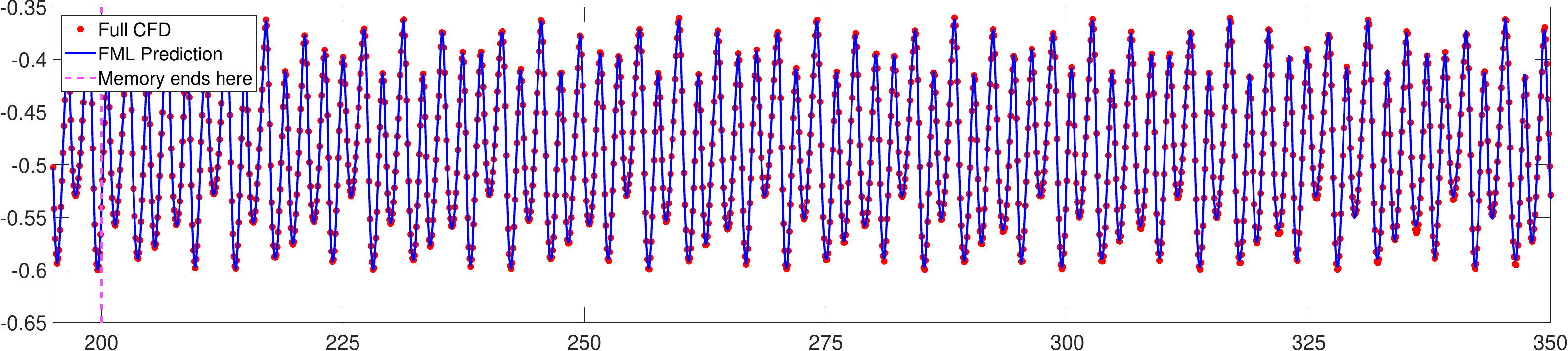} 
    \caption{Coefficient $a_1$}
    \end{subfigure} 
\begin{subfigure}[t]{0.9\textwidth}
\includegraphics[width=\textwidth,]{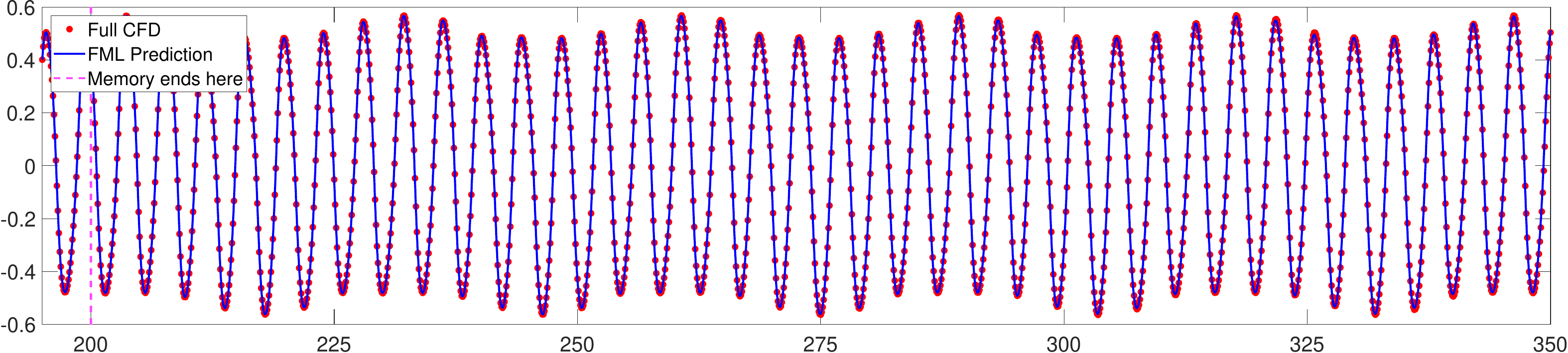} 
    \caption{Coefficient $b_1$}
    \end{subfigure}  
    \begin{subfigure}[t]{0.9\textwidth}
\includegraphics[width=\textwidth,]{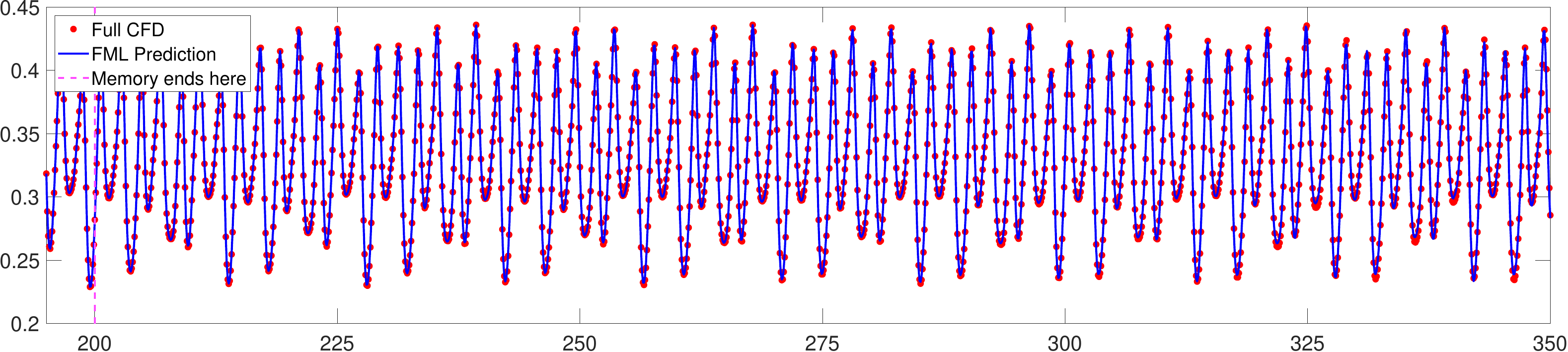} 
    \caption{Coefficient $a_2$}
    \end{subfigure}   
    \begin{subfigure}[t]{0.9\textwidth}
\includegraphics[width=\textwidth,]{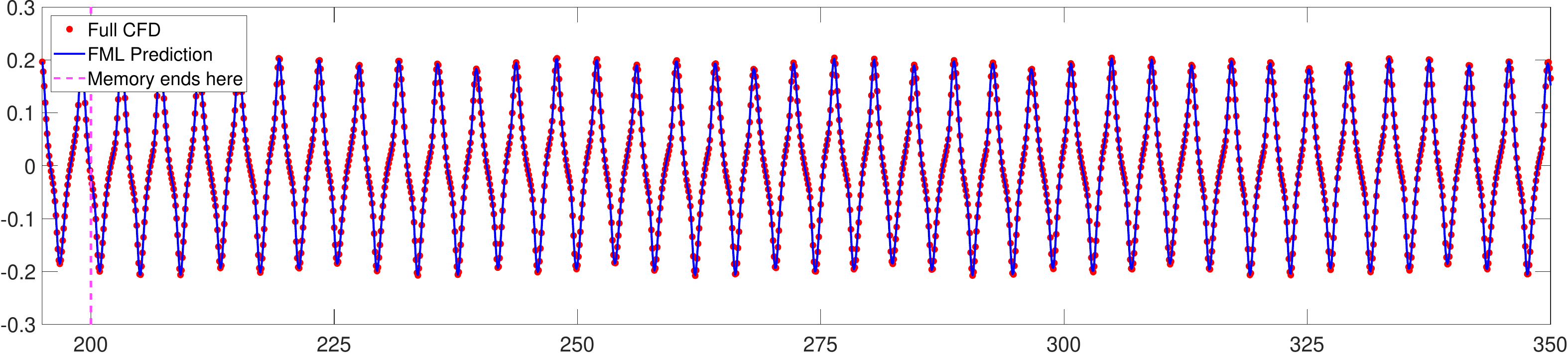} 
    \caption{Coefficient $b_2$}
    \end{subfigure}  
		\caption{Test 2-1: Comparison of the Fourier coefficients evolution between the tDT-2 model prediction and the full CFD DT reference solution at $Re=1,500$ for $\hat{t}\leq 350$.}
\label{fig:fourier}
	\end{center}
\end{figure}

In Figure~\ref{fig:fourier}, we show the temporal evolutions of the first few dominating Fourier coefficients for up to $\hat{t}=350.0$, which corresponds to tDT-2 predictions up to $t=150.0$. (We choose not to display the full prediction up to $\hat{t}=485.0$ because the figures would become too cluttered.) We observe good agreements between the tDT-2 predictions and the reference solutions by the CFD full DT simulation.

We then reconstruct the pressure distribution on the cylinder surface via \eqref{Ps} and examine its evolution over time. The snapshots of the pressure distribution over polar angle are shown in Figure~\ref{fig:P_Re1500}, where $\theta=0$ is the rear end of the cylinder and $\theta=\pm\pi$ is the front end. The surface pressure is plotted over every 15 time units within the prediction time frame $200\leq \hat{t}\leq 485$. We observe that the tDT-2 model is capable of predicting and capturing the fine details of the surface pressure function and provides good long-term accuracy. 

In a different test case (Test 2-2), we conducted tDT-2 prediction using initial conditions obtained from the full DT result at $Re=1,000$ and between $195.0\leq \hat{t}\leq 200.0$. The predictions of the surface pressure for up to $\hat{t}=485.0$ are plotted in Figure~\ref{fig:P_Re1000}. Again, we observe very good predictive accuracy of the tDT-2 model. 

\begin{figure}[htbp]
	\begin{subfigure}[t]{0.48\textwidth}
            \centering
            \includegraphics[width=1\textwidth]{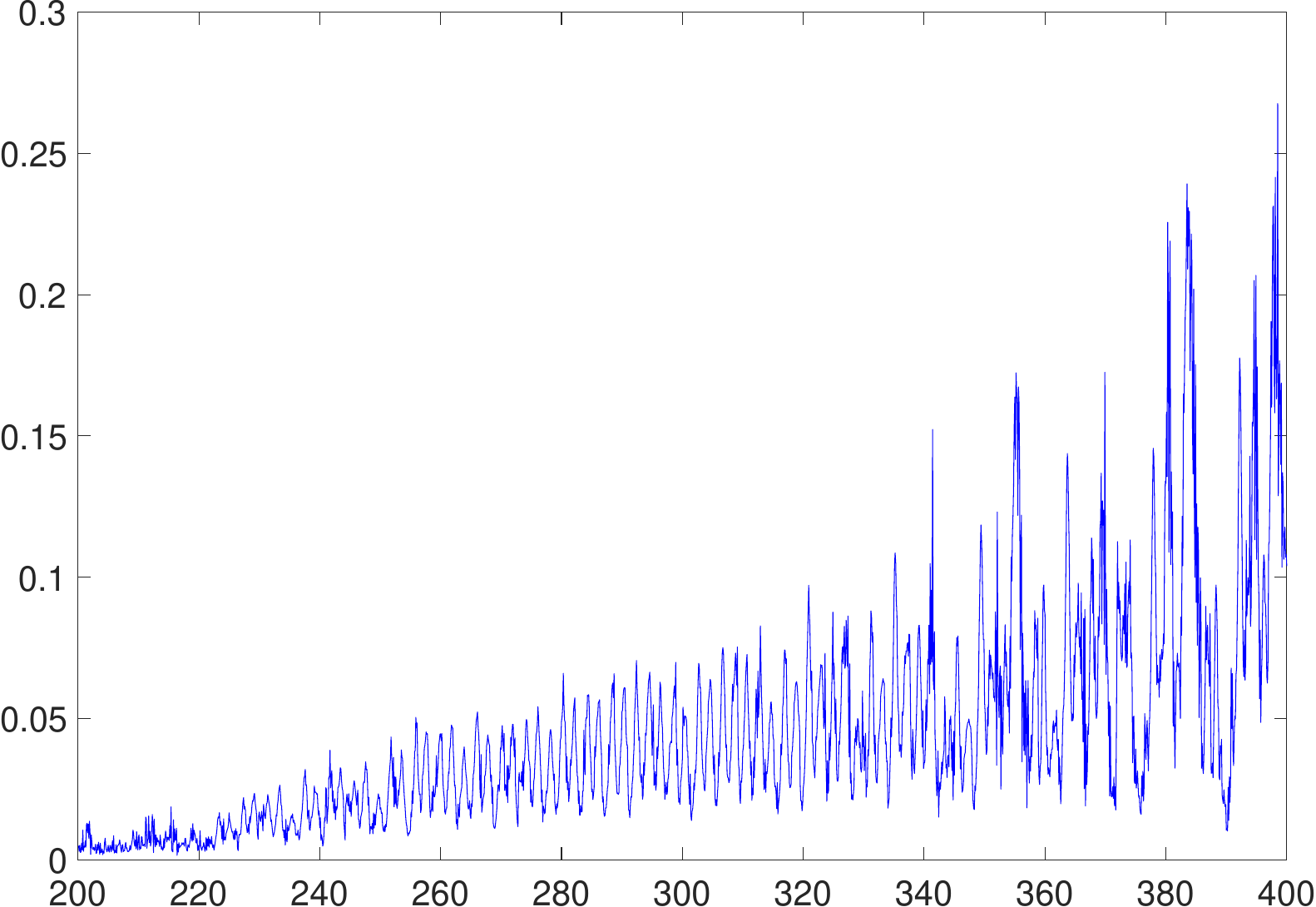}
        \end{subfigure}
        \begin{subfigure}[t]{0.48\textwidth}
            \centering
            \includegraphics[width=1\textwidth]{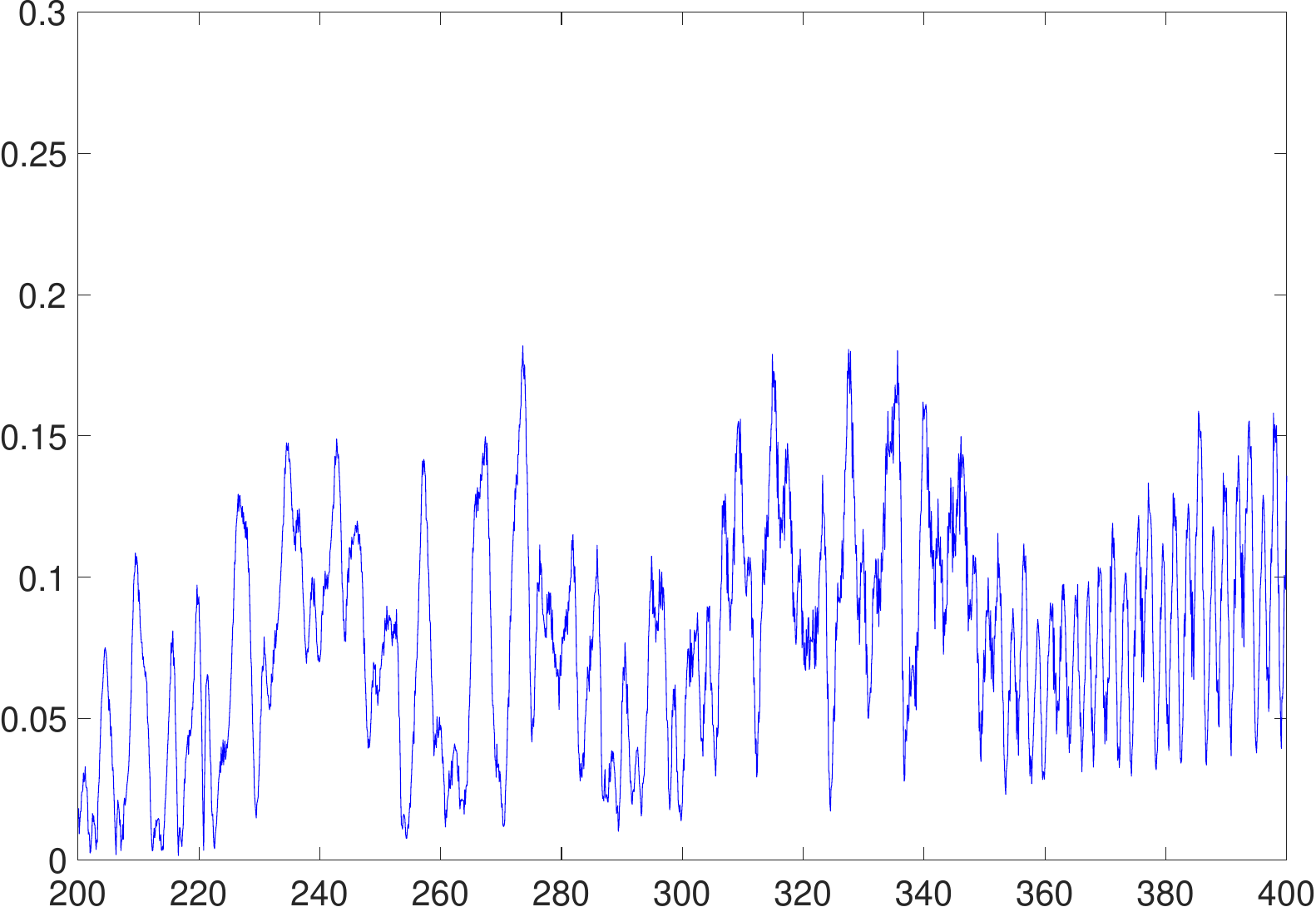}
        \end{subfigure}
		\caption{Evolution of $L^2$ error in tDT prediction of the surface pressure. Left: Test 2-1 with $Re=1,500$. Right: Test 2-2 with $Re=1,000$.}
\label{fig:P_error}
\end{figure}

The $L^2$ errors of the surface pressure function $P_S$ predicted by the tDT-2 model, against the full DT reference solution $p_{ref}$, is defined as
$$
e_2(t) = \left(\int_{-\pi}^{\pi} \left(P_s(t,\theta) - p_{ref}(t,\theta)\right)^2d\theta\right)^{1/2}.
$$
The errors are plotted in Figure~\ref{fig:P_error}, for physical time up to $\hat{t}\leq 400.0$ for both test cases. Again, we observe a relatively small growth of the errors over the long-term predictions. Such a growth is not unexpected due to the quai-periodic nature of the solution evolution. 

Many other test cases were examined at different $Re$ numbers and different initial conditions in physical time. The results are similar to the two cases presented here.

\begin{figure}[!ht]
    \centering
    \foreach \n in {1,...,20}{
    \pgfmathtruncatemacro{\captionnum}{185+15*\n}
        \begin{subfigure}[b]{0.18\textwidth}
            \centering                  
            \includegraphics[width=\linewidth]{re1500pressure_time\captionnum-eps-converted-to.pdf}
            \caption{Time \captionnum}
        \end{subfigure}%
        \pgfmathparse{mod(\n, 5) == 0 ? 1 : 0}
        \ifnum\pgfmathresult=1
            \ifnum\n<20
                \\[\medskipamount] 
            \fi
        \else
            \hfill 
        \fi
    }
    \caption{Test 2-1: Comparison of snapshots of the surface pressure distributions by the tDT-2 model prediction and the full CFD DT reference solutions ($Re=1,500$).}
    \label{fig:P_Re1500}
\end{figure}

\begin{figure}[!ht]
    \centering
    \foreach \n in {1,...,20}{
    \pgfmathtruncatemacro{\captionnum}{185+15*\n}
        \begin{subfigure}[b]{0.18\textwidth}
            \centering                  
            \includegraphics[width=\linewidth]{re1000pressure_time\captionnum-eps-converted-to.pdf}
            \caption{Time \captionnum}
        \end{subfigure}%
        \pgfmathparse{mod(\n, 5) == 0 ? 1 : 0}
        \ifnum\pgfmathresult=1
            \ifnum\n<20
                \\[\medskipamount] 
            \fi
        \else
            \hfill 
        \fi
    }
    \caption{Test 2-2: Comparison of snapshots of the surface pressure distributions by the tDT-2 model prediction and the full CFD DT reference solutions ($Re=1,000$).}
    \label{fig:P_Re1000}
\end{figure}

\section{Conclusion} \label{sec:conclusions}

We presented in this paper a numerical procedure for constructing
Targeted Digital Twin (tDT), which is defined as a direct numerical
modeling of the quantities-of-interest (QoIs) of a full digital
twin. Relying on FML (flow map learning) methodology, we discussed the
dynamical model of the tDT, the generation of training data by
repeatedly executing the full DT, and the training of the tDT. To
demonstrate the efficacy of the approach, we employed the well studied
CFD problem of two-dimensional flow past a cylinder. We demonstrated
that through expensive ``offline" computations, which include training
data generation and tDT model learning, the tDT provides a highly
efficient way for system analysis and prediction of the QoIs without
requiring to the full DT. 
This significantly speeds up the online QoI
computations. Consequently, the tDT approach offers a possibility for
real-time digital twin implementation.

\bibliographystyle{siamplain}
\bibliography{neural, LearningEqs, ensemble, DT}

\end{document}